\documentclass{article}
\usepackage[utf8]{inputenc}
\usepackage{main}
\usepackage{microtype}
\usepackage{graphicx}
\usepackage{times}
\usepackage{amsmath}
\usepackage{amssymb}
\usepackage{enumitem}
\usepackage{booktabs}
\usepackage{xcolor}     
\usepackage{natbib}
\usepackage{forest}
\definecolor{mydarkblue}{rgb}{0,0.08,0.45}
\usepackage[colorlinks=true,linkcolor=mydarkblue,citecolor=mydarkblue,filecolor=mydarkblue,urlcolor=mydarkblue]{hyperref}
\usepackage{fancyhdr}

\definecolor{myblue}{RGB}{230,242,255}
\definecolor{mypurple}{RGB}{238,230,255}
\definecolor{myred}{RGB}{255,235,238}
\definecolor{mygreen}{RGB}{232,245,233}
\definecolor{myteal}{RGB}{224,242,241}

\newcommand{\github}{\raisebox{-1.5pt}{\includegraphics[height=1.05em]{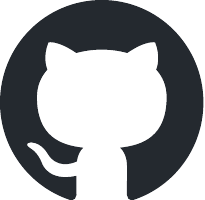}}\xspace}

\definecolor{gred}{RGB}{250, 210, 207}
\definecolor{coolblue1}{rgb}{0.91, 0.94, 0.98}
\definecolor{coolblue2}{rgb}{0.76, 0.85, 0.94}
\definecolor{coolblue3}{rgb}{0.54, 0.72, 0.87}
\definecolor{coolblue4}{rgb}{1, 1, 1}

\usepackage[table]{xcolor}  

\usepackage{xspace}
\usepackage{cleveref}
\usepackage[]{multicol, multirow}
\usepackage[most]{tcolorbox}
\usepackage{etoc}

\tcbuselibrary{listingsutf8}
\tcbuselibrary{listingsutf8,breakable}

\newtcolorbox[auto counter]{observation}[1][]{
  colback=black!5!white,
  colframe=black!70!white,
  fonttitle=\bfseries,
  title=Observation~\thetcbcounter,
  enhanced,
  boxrule=0.6pt,
  left=1mm,right=1mm,top=1mm,bottom=1mm,
  #1
}

\newtcolorbox[auto counter]{takeaway}[1][]{
  colback=teal!3!white,
  colframe=teal!55!black,
  fonttitle=\bfseries,
  title=Takeaway~\thetcbcounter,
  enhanced,
  boxrule=0.5pt,
  left=1mm,right=1mm,top=1mm,bottom=1mm,
  #1
}

\newtcolorbox[auto counter]{practicalguidance}[1][]{
  colback=cyan!3!white,
  colframe=cyan!60!black,
  fonttitle=\bfseries,
  title=Practical Guidance~\thetcbcounter,
  enhanced,
  boxrule=0.5pt,
  left=1mm,right=1mm,top=1mm,bottom=1mm,
  #1
}

\newtcolorbox[auto counter]{discussion}[1][]{
  colback=violet!4!white,
  colframe=violet!60!black,
  fonttitle=\bfseries,
  title=Discussion~\thetcbcounter,
  enhanced,
  boxrule=0.5pt,
  left=1mm,right=1mm,top=1mm,bottom=1mm,
  #1
}

\newenvironment{itemize*}%
 {\leftmargini=10pt\begin{itemize}%
  \setlength{\itemsep}{0pt}%
  \setlength{\parskip}{0pt}%
  }%
 {\end{itemize}}
\newenvironment{enumerate*}%
 {\begin{enumerate}%
  \setlength{\itemsep}{0pt}%
  \setlength{\parskip}{0pt}}%
 {\end{enumerate}}

\title{A Survey of Multimodal Mathematical Reasoning: From Perception,
Alignment to Reasoning}

\author{
Tianyu Yang$^{1\thanks{Equal contributions.}}$, \xspace
Sihong Wu$^{2\footnotemark[1]}$, \xspace
Yilun Zhao$^{2\footnotemark[1]}$, \xspace
Zhenwen Liang$^{1}$, 
Lisen Dai$^{3}$,
\vspace{2pt} \\
\bf{
Chen Zhao$^{4}$,
Minhao Cheng$^{5}$,
Arman Cohan$^{2}$,
Xiangliang Zhang$^{1}$\thanks{Correspondence} \vspace{2pt}}\\
$^{1}$University of Notre Dame \quad
$^{2}$Yale University \quad
$^{3}$Columbia University \quad \vspace{3pt}\\
$^{4}$New York University \quad
$^{5}$Pennsylvania State University \vspace{3pt}
\\
\texttt{\{tyang4, xzhang33\}@nd.edu}
}

\begin{document}

\maketitle
\thispagestyle{fancy}
\fancyhead{}
\lhead{
\includegraphics[height=1.3cm]{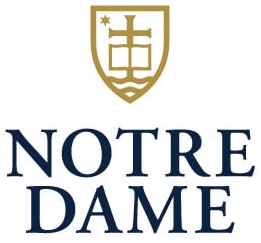}\hspace{0.2cm}
\includegraphics[height=1cm]{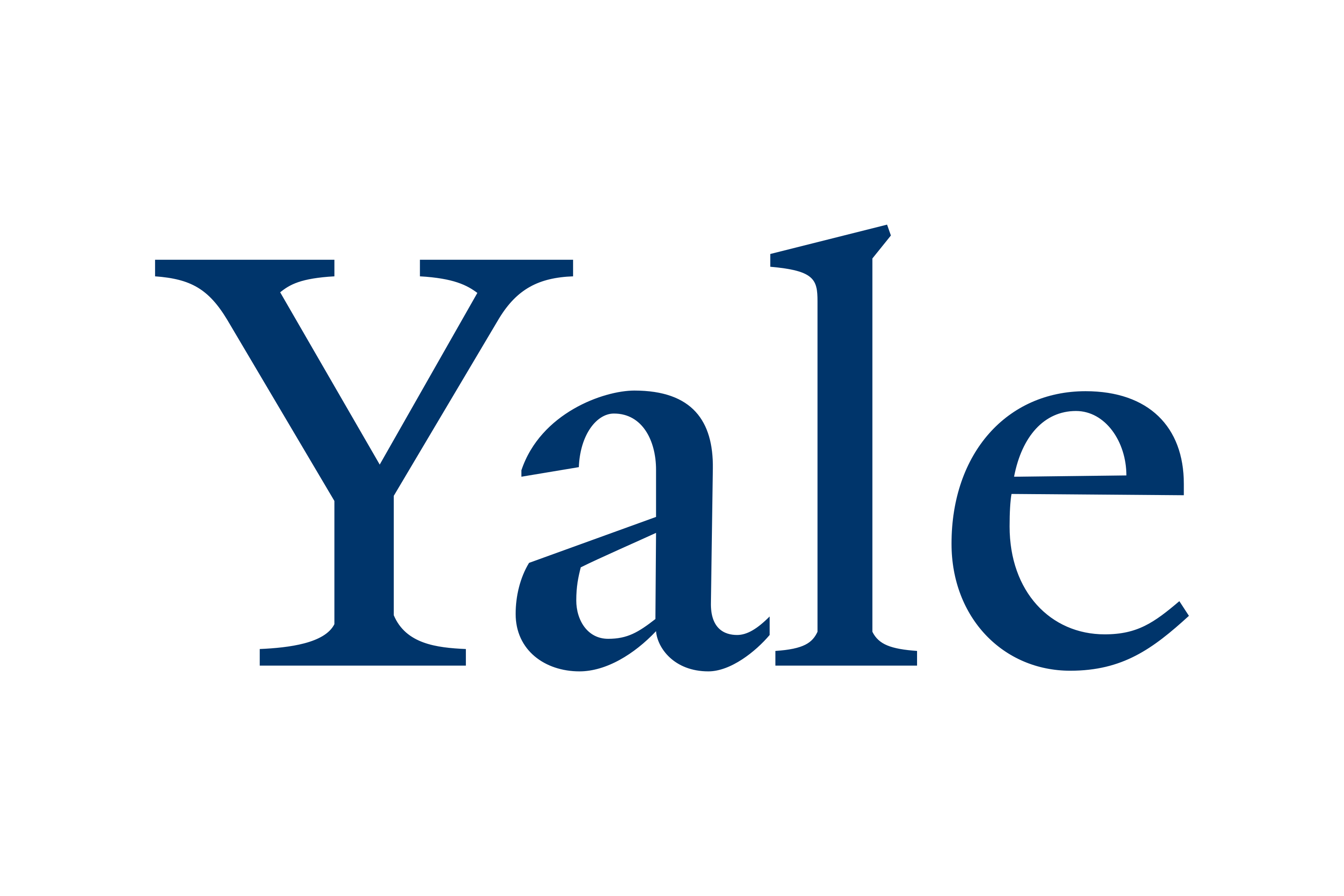}\hspace{0.2cm}
\includegraphics[height=1.3cm]{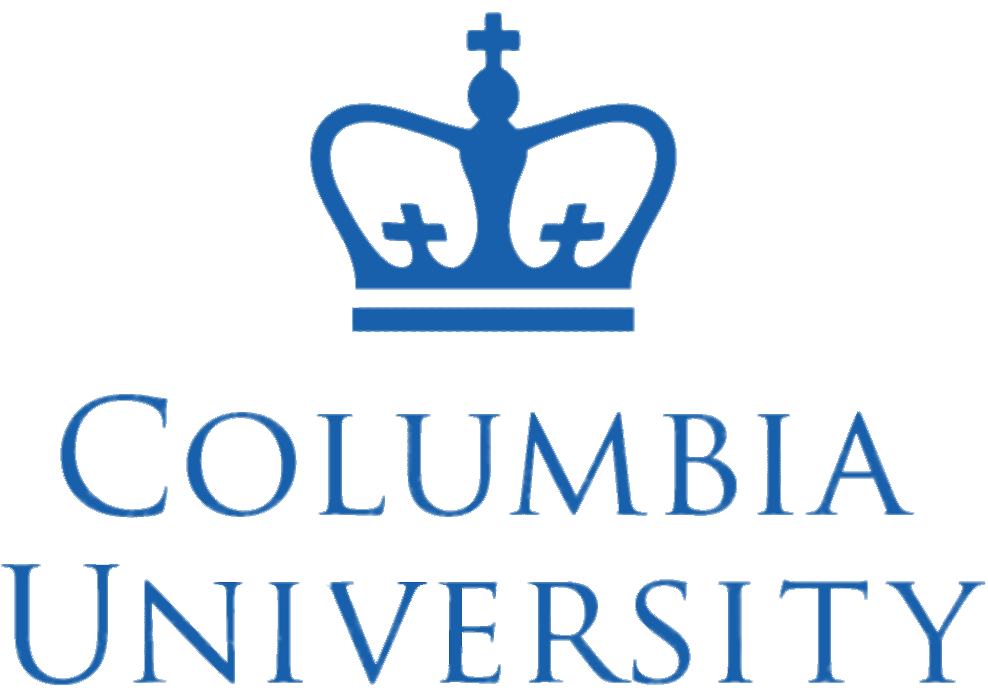}\hspace{0.4cm}
\includegraphics[height=1.3cm]{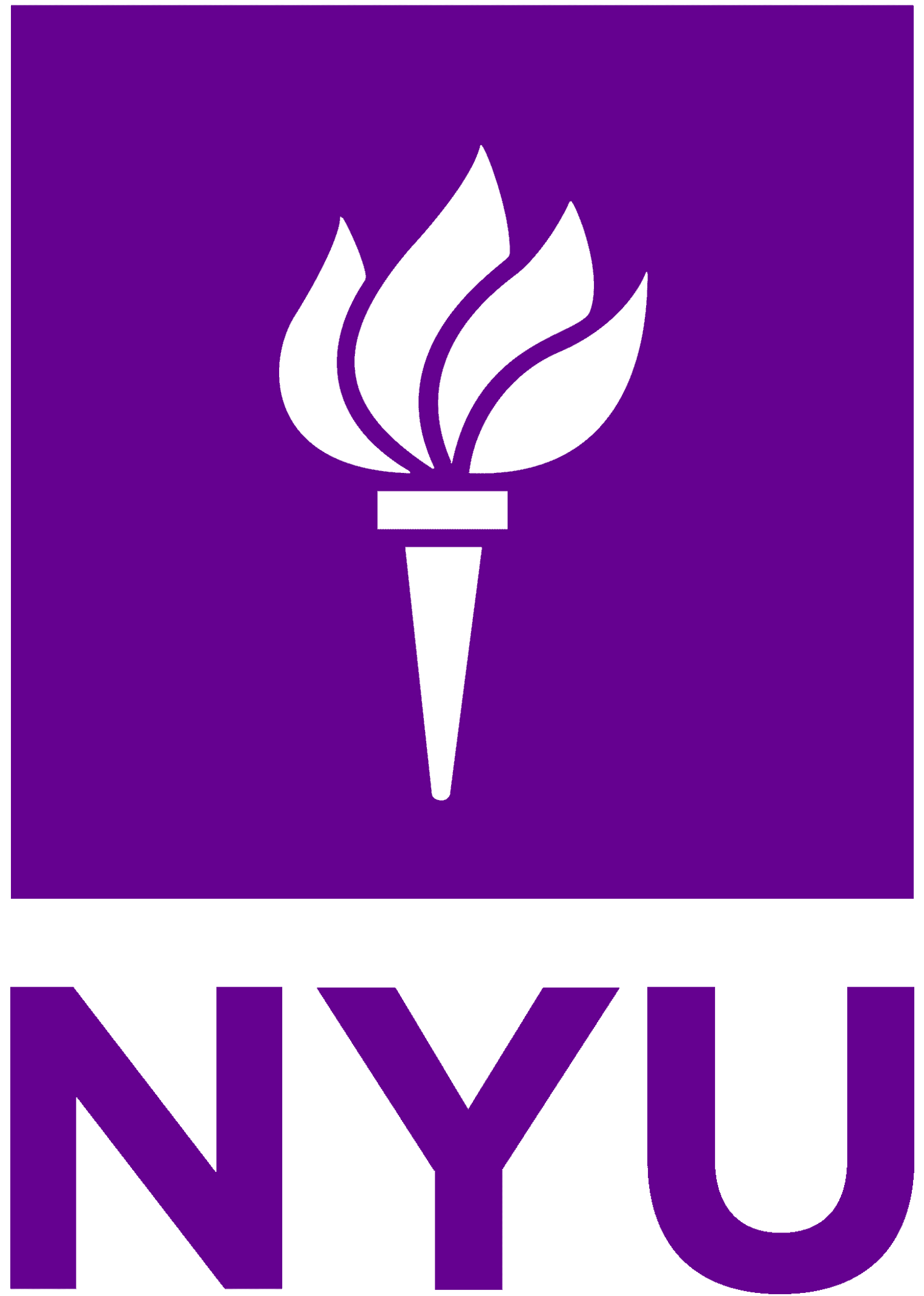}\hspace{0.3cm}
\includegraphics[height=1.4cm]{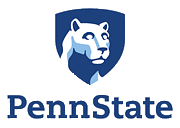}\hspace{0.2cm}
}
\renewcommand{\headrulewidth}{0pt}
\setlength{\headheight}{30pt}
\addtolength{\topmargin}{00pt}
\setlength{\headsep}{3mm}

\begin{abstract}
Multimodal Mathematical Reasoning (MMR) has recently attracted increasing attention for its capability to solve mathematical problems involving both textual and visual modalities. However, current models still face significant challenges in real-world visual math tasks, often misinterpreting diagrams, failing to align mathematical symbols with visual evidence, or producing inconsistent reasoning steps. Moreover, existing evaluations mainly focus on checking final answers rather than verifying the correctness or executability of each intermediate step. A growing body of recent research addresses these issues by integrating structured perception, explicit alignment, and verifiable reasoning within unified frameworks.
To establish a clear roadmap for understanding and comparing different MMR approaches, we systematically review them around four fundamental questions: (1) What to extract from multimodal inputs, (2) How to represent and align textual and visual information, (3) How to perform the reasoning, and (4) How to evaluate the correctness of the overall reasoning process. Finally, we discuss open challenges and share our thoughts on future research directions.
\end{abstract}

\begin{center}
\begin{tabular}{c}
\github\ \href{https://github.com/formula12/Awesome-Multimodal-Mathematical-Reasoning-Perception-Alignment-Reasoning}{\textbf{GitHub:} Awesome Multimodal Mathematical Reasoning}
\end{tabular}
\end{center}
\vspace{5pt}

\begin{figure}[h]
\centering
\includegraphics[width=0.5\textwidth]{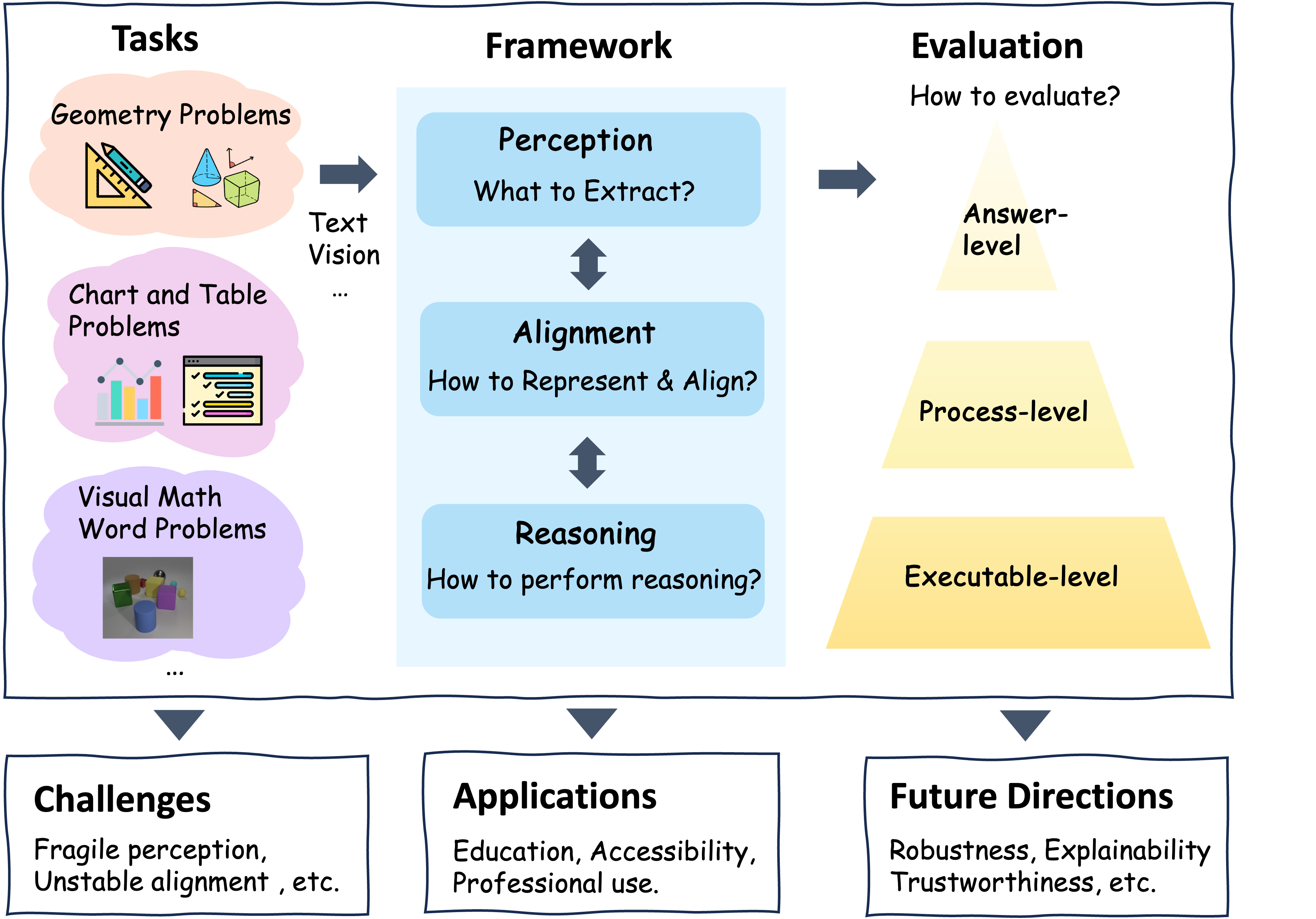}
\caption{
The roadmap of this survey.
}
\label{intro}
\end{figure}

\section{Introduction}

\begin{figure*}[ht]
\centering
\resizebox{\textwidth}{!}{
\begin{forest}
  for tree={
    grow=east,
    reversed=true,
    anchor=base west,
    parent anchor=east,
    child anchor=west,
    base=middle,
    font=\small,
    rectangle,
    draw=black,
    edge=black!50,
    edge path'={%
        (!u.parent anchor) -- +(.9em,0) |- (.child anchor)
      },
    edge+={line cap=round},
    align=center,
    minimum width=1em,
    s sep=6pt,
    inner xsep=2.5pt,
    inner ysep=1.2pt,
    text centered,
  },
    where level=0{text width=25em,font=\normalsize,}{},
    where level=1{text width=8em,font=\normalsize,}{},
    where level=2{text width=8em,font=\normalsize,}{},
    where level=3{text width=55em,font=\normalsize,}{},
    where level=4{text width=43.3em,font=\normalsize,}{},
  [Perception–Alignment–Reasoning (PAR) framwork,parent anchor=south, child anchor=west,rotate=90,anchor=north,edge=black!50,fill=myblue,draw=black
    [What to Extract?,edge=black!50, fill=mypurple, minimum height=1.2em
      [Geometry,fill=mypurple
        [{GEOS \cite{seo-etal-2015-solving};
        E-GPS \cite{wu2024gps};
        Pi-GPS \cite{zhao2025pi};
        GeomVerse \cite{kazemi2023geomversesystematicevaluationlarge};\\
         G-LLaVA \cite{gao2023gllavasolvinggeometricproblem}; 
        GeoGPT4V \cite{cai2024geogpt4vgeometricmultimodallarge}; GeoQA+ \cite{cao-xiao-2022-augmented};
        Geometry3K \cite{lu-etal-2021-inter}; \\ GeoQA \cite{chen2022geoqageometricquestionanswering}; 
        PGDP5K \cite{hao2022pgdp5kdiagramparsingdataset}; PGPS9K \cite{zhang2023multimodalneuralgeometricsolver};\\
        DFE-GPS \cite{xin2025generalizable}; 
        GEOX \cite{xia2024geox}},
         fill=mypurple]
      ]
      [Chart and Table, fill=mypurple
        [{PlotQA \cite{methani2020plotqareasoningscientificplots}; ChartQA \cite{masry2022chartqa}; FinQA \cite{chen2022finqadatasetnumericalreasoning}; TATQA \cite{zhu2021tatqaquestionansweringbenchmark}; \\
        MultiHiertt~\cite{zhao-etal-2022-multihiertt}; DocMath-Eval~\cite{zhao-etal-2024-docmath}; 
        DVQA \cite{kafle2018dvqa}; PlotQA \cite{methani2020plotqa}; \\ ChartQA \cite{masry2022chartqa}; Pix2Struct \cite{lee2023pix2struct}; 
        Chartx \cite{xia2024chartx}; Chartllama \cite{han2023chartllama}; \\DePlot \cite{liu2023deplotoneshotvisuallanguage}; LogicNLG~\cite{chen-etal-2020-logical}},
         fill=mypurple]
      ]
      [Visual Math\\Word Problems, fill=mypurple
        [{IconQA \cite{lu2021iconqa}; CLEVR Math \cite{lindstrom2022clevr}; MV MATH \cite{wang2025mv}; \\
        Patch-TRM \cite{lu2021iconqa}; Geogpt4v \cite{cai2024geogpt4v}; Inter-gps \cite{lu2021inter}; TABMWP~\cite{lu2022dynamic}},
         fill=mypurple]
      ]
    ]
    [How to Align?, edge=black!50, fill=myred, minimum height=1.2em
      [Executable\\Intermediates, fill=myred
        [{GeoQA+ \cite{cao2022augmented}; FormalGeo \cite{zhang2024formalgeoextensibleformalizedframework};\\
         Inter GPS \cite{lu2021inter}; E-GPS \cite{wu2024gps}; Pi GPS \cite{zhao2025pi}},
         fill=myred]
      ]
      [Symbolic--Neural\\Hybrids, fill=myred
        [{GeoGen \cite{pan2025enhancing}; AlphaGeometry \cite{trinh2024solving}; MathCoder VL \cite{wang2025mathcoder}},
         fill=myred]
      ]
      [Cross-modal\\Alignment, fill=myred
        [{BLIP 2 \cite{li2023blip}; LLaVA \cite{liu2023visual}; Math PUMA \cite{zhuang2024mathpumaprogressiveupwardmultimodal}; 
        VCAR \cite{jia2024describe}; \\TVC \cite{sun2025mitigating}; VIC \cite{zheng2024thinking}},
         fill=myred]
      ]
      [Pre-training\&\\Fine-tuning, fill=myred
        [{GeoGen \cite{pan2025enhancing}; MathCoder VL \cite{wang2025mathcoder}; SynthGeo228K \cite{zhang2025diagram}; \\
        GeoGPT-4V \cite{cai2024geogpt4v}; Math-LLaVA \cite{shi2024math}; MAVIS \cite{zhang2024mavis}; \\ MultiMath-300K \cite{peng2024multimath}; 
        AlphaGeometry \cite{trinh2024solving}; AtomThink \cite{xiang2024atomthink}; \\Masked Thought \cite{chen2024masked}; 
         LogicSolver \cite{yang2022logicsolver}; MathGenie \cite{lu2024mathgenie}; \\MMathCoT-1M \cite{shi2024math};
         DualMath-1.1M \cite{zhang2024mavis}; MathV360K \cite{shi2024math}; Inter-GPS \cite{lu2021inter};\\
         GeoQA\cite{chen2021geoqa}; GeoQA+ \cite{cao2022augmented}; E-GPS \cite{wu2024gps}; VCAR \cite{jia2024describe}; \\
         Math-PUMA \cite{zhuang2024mathpumaprogressiveupwardmultimodal}; MAmmoTH-VL \cite{guo2024mammoth}; TrustGeoGen \cite{fu2025trustgeogen}} ,fill=myred]
      ]
    ]
    [How to Perform\\Reasoning?, edge=black!50, fill=mygreen, minimum height=1.2em
      [Deliberate Chains, fill=mygreen
        [{LLaVA CoT \cite{xu2024llava}; VisuoThink \cite{wang2025visuothink}; VReST \cite{zhang2025vrest}; ToT \cite{yao2023tree};\\
         GoT \cite{besta2024graph}},
         fill=mygreen]
      ]
      [RL-based, fill=mygreen
        [Reward Mechanism\\Design, fill=mygreen, text width=10em
          [{R1 VL \cite{zhang2025r1}; VisualPRM \cite{wang2025visualprm};\\ MM PRM \cite{du2025mm}; 
           MM Eureka \cite{meng2025mm}},
           fill=mygreen]
        ]
        [Search \& Decision\\Algorithms, fill=mygreen, text width=10em
          [{DeepSeek-R1 \cite{guo2025deepseek}; Vision R1 \cite{huang2025vision}; Mulberry \cite{yao2412mulberry}; \\
          Skywork R1V2 \cite{chris2504skywork};  VL Rethinker \cite{wang2025vl}; FAST \cite{xiao2025fast}; \\
           AlphaProof~\cite{deepmind2024alphaproof}; 
           Think or Not \cite{wang2025think};  VLAA Thinking \cite{chen2025sft};\\ 
           VLM R3 \cite{jiang2025vlm}; MAYE \cite{ma2025rethinking}; SoTA with Less \cite{wang2025sota}},
           fill=mygreen]
        ]
      ]
      [Process Feedback\\\& Verification, fill=mygreen
        [{VisualPRM \cite{wang2025visualprm}; MM-PRM \cite{du2025mm}; TVC \cite{sun2025mitigating}; 
        VIC \cite{zheng2024thinking}},
         fill=mygreen]
      ]
      [Tool Augmented, fill=mygreen
        [{MathCoder-VL \cite{wang2025mathcoder}; Pi-GPS \cite{zhao2025pi}; Visual Sketchpad \cite{hu2024visual}; \\
        Toolformer \cite{schick2023toolformer}; ToRA \cite{gou2023tora}; MM REACT \cite{yang2023mm}},
         fill=mygreen]
      ]
      [Supervision \& Data, fill=mygreen
        [Error Detection\\and Correction, fill=mygreen, text width=10em
          [{MM MATH \cite{sun2024mm}; ErrorRadar \cite{yan2024errorradar}; MPBench \cite{xu2025mpbench}; \\
          Sherlock \cite{ding2025sherlock}; We Math \cite{qiao2024we}; \\Mathador LM \cite{kurtic2024mathador};
           VATE \cite{xu2025ai}},
           fill=mygreen]
        ]
        [Mathematical Problem\\Generation, fill=mygreen, text width=10em
          [{GeoGen \cite{pan2025enhancing}; GeoGPT 4V \cite{cai2024geogpt4v}; Math LLaVA \cite{shi2024math}; \\
          MAVIS \cite{zhang2024mavis}; MultiMath 300K \cite{peng2024multimath}; AtomThink \cite{xiang2024atomthink}},
           fill=mygreen]
        ]
      ]
    ]
  ]
\end{forest}
}

\vspace{-0.03in}
\caption{Taxonomy of Perception, Alignment and Reasoning framework.} \vspace{-0.1in}
\label{fig:taxonomy}
\end{figure*}

Large Language Models (LLMs) have recently advanced mathematical reasoning, achieving state-of-the-art results on various symbolic and arithmetic tasks, from elementary school level to college level \cite{deepmind2024alphaproof, guo2025deepseek}. However, 
in practice, mathematics often involves multimodal information. 
Many real-world problems in education \cite{ku2025theoremexplainagent}, scientific discovery \cite{du2025mm}, and interactive professional systems \cite{hu2024visual, zhao2025mmvu} require reasoning over visual structures and spatial relations. Solving these problems often requires interpreting diagrams, coordinate plots, charts, tables, and mixed-modality documents \cite{lu2021iconqa, saikh2022scienceqa, lee2023pix2struct, zhao-etal-2023-qtsumm}. In these contexts, visual elements encode critical constraints—such as incidence, parallelism, numeric scales, and layout semantics—that text-only models simply cannot perceive \cite{chen2025spatial}.

To handle this complexity, a line of work focuses on integrating perception, symbolic understanding, and executable reasoning across modalities, defining the field of Multimodal Mathematical Reasoning (MMR) \cite{chen2021geoqa, lu2021iconqa, saikh2022scienceqa}. Compared with purely text-based approaches \cite{lewkowycz2022solving, liang2023unimath}, MMR approaches significantly improves evidence completeness by grounding visual cues. Nonetheless, these multimodal learning approaches substantially increase reasoning complexity: a model must jointly interpret visual cues, align them with symbolic expressions, and execute consistent multi-step reasoning across modalities \cite{chen2021geoqa, sheng2025solving}. This strong multimodal coupling introduces new, non-trivial challenges related to structured perception, cross-modal alignment, and verifiable reasoning. .

Given the importance of MMR and its rapid progress, we are motivated to present this survey that foregrounds fundamental mechanisms of addressing MMR using Multimodal LLMs (MLLMs). 
Prior efforts  primarily catalog benchmarks and methodologies for MMR \cite{yan2024survey} or  discuss MLLM ecosystem roles (Reasoner, Enhancer, Planner) \cite{yan2024survey}. 
In contrast, we take a \textbf{vertical, process-centric view}: we articulate what is needed to solve MMR end-to-end and position MLLM-based approaches along this roadmap.
Concretely, we organize the field around four questions: \textbf{1)}  what to extract from multimodal inputs, \textbf{2)} how to represent and align textual and visual information, \textbf{3)} how to perform the reasoning (e.g., CoT, program-aided, tool use), and \textbf{4)} how to evaluate the correctness of the reasoning process.
More discussion about our work vs related surveys is provided in Table~\ref{tab:Comparisons} and Appendix $\ref{Related Surveys}$.

\begin{table*}[t]
\centering
\tiny
\setlength{\tabcolsep}{4pt}
\resizebox{1\textwidth}{!}{%
\begin{tabular}{l l l l p{3.4cm}}
\toprule
\textbf{Benchmark} & \textbf{Year (Venue)} & \textbf{Eval Level} & \textbf{PAR Stage} & \textbf{Key Contributions} \\
\midrule
ChartQA~\cite{masry2022chartqa} & 2022 (ACL Findings) & Answer & Perception + Reasoning & Real charts; logical \& numeric QA. \\
FigureQA~\cite{kahou2017figureqa} & 2018 (ICLR Workshop) & Answer & Perception & Synthetic charts; controlled reasoning. \\
PlotQA~\cite{methani2020plotqareasoningscientificplots} & 2020 (WACV) & Answer & Perception + Reasoning & Real plots; open‑vocab numeric answers. \\
IconQA~\cite{lu2021iconqa} & 2021 (NeurIPS) & Answer & Perception + Reasoning & Large icon‑based multimodal math. \\
CLEVR‑Math~\cite{lindstrom2022clevr} & 2022 (NeSy Workshop) & Answer & Perception + Reasoning & Synthetic compositional arithmetic. \\
FinQA~\cite{chen2022finqadatasetnumericalreasoning} & 2021 (EMNLP) & Answer & Alignment + Reasoning & Financial table‑text; gold programs. \\
TAT‑QA~\cite{zhu2021tatqaquestionansweringbenchmark} & 2021 (ACL) & Answer & Alignment + Reasoning & Table‑text numeracy in reports. \\
MultiHiertt~\cite{zhao-etal-2022-multihiertt} & 2022 (ACL) & Answer & Alignment + Reasoning & Financial table‑text; gold programs. \\
DocMath-Eval~\cite{zhao-etal-2024-docmath} & 2024 (ACL) & Answer & Alignment + Reasoning & Financial table‑text; gold evidence. \\
 ChartQAPro~\cite{masry-etal-2025-chartqapro} & 2025 (ACL Findings) & Answer & Perception + Alignment & Harder charts incl. dashboards. \\
 CharXiv~\cite{wang2024charxiv} & 2024 (NeurIPS) & Answer & Perception & Human‑curated arXiv charts. \\
\midrule
MM‑MATH~\cite{sun2024mm} & 2024 (EMNLP Findings) & Process & Reasoning & Step types \& error labels. \\
MPBench~\cite{xu2025mpbench} & 2025 (ACL Findings) & Process & Reasoning & PRM / step‑judge benchmarking. \\
ErrorRadar~\cite{yan2024errorradar} & 2024 (ICLR Workshop) & Process & Reasoning & Fine‑grained error taxonomy. \\
Sherlock~\cite{ding2025sherlock} & 2025 (NeurIPS) & Process & Reasoning & Multimodal error detect \& repair. \\
We‑Math~\cite{qiao2024we} & 2025 (ACL) & Process & Reasoning & Principle‑centered process probing. \\
MathVerse~\cite{zhang2024mathverse} & 2024 (ECCV) & Process & All & Diagram perturbations; CoT step scoring. \\
CHAMP~\cite{mao2024champ} & 2024 (ACL Findings) & Process & Reasoning & Competition items; wrong‑step tags. \\
PolyMATH~\cite{gupta2024polymath} & 2024 (arXiv) & Process & Perception + Reasoning & Image–text puzzles; cognitive coverage. \\
\midrule
GeoQA+~\cite{cao2022augmented} & 2022 (COLING) & Executable & Alignment + Reasoning & Geometry QA with executable programs. \\
Geometry3K~\cite{lu-etal-2021-inter} & 2021 (ACL) & Executable & Perception + Alignment & Dense formal language for geometry. \\
E‑GPS~\cite{lu2021inter,wu2024gps} & 2024 (CVPR) & Executable & All & Solver+parser; verifiable steps. \\
FormalGeo~\cite{zhang2024formalgeoextensibleformalizedframework} & 2024 (MATH‑AI) & Executable & Alignment + Reasoning & Olympiad‑level formal proofs. \\
Pi‑GPS~\cite{zhao2025pi} & 2025 (arXiv) & Executable & Alignment + Reasoning & Rectifier and solver for proofs. \\
WikiSQL~\cite{zhong2017seq2sqlgeneratingstructuredqueries} & 2017 (arxiv) & Executable & Alignment + Reasoning & NL→SQL with execution accuracy. \\
\midrule
MathVista~\cite{lu2024mathvistaevaluatingmathematicalreasoning} & 2024 (ICLR) & Comprehensive & All & Aggregated multimodal suite. \\
MATH‑V~\cite{wang2024measuring} & 2024 (NeurIPS) & Comprehensive & All & Difficulty‑calibrated visual math. \\
OlympiadBench~\cite{he2024olympiadbench} & 2024 (ACL) & Comprehensive & All & Bilingual competition‑grade; stepwise. \\
MathScape~\cite{liang2024mathscape} & 2024 (arXiv) & Comprehensive & All & Photo scenarios; multi‑dim evaluation. \\
CMM-Math~\cite{liu2024cmm} & 2024 (ACMMM) & Comprehensive & All & Chinese multimodal math. \\
Children’s Olympiads~\cite{cherian2024evaluating} & 2024 (ESEM) & Comprehensive & All & Olympiad-style problems. \\
MM-PRM~\cite{du2025mm} & 2025 (arXiv) & Comprehensive & All & Real-world K-12 multimodal QA. \\
\bottomrule
\end{tabular}%
}\vspace{-0.05in}
\caption{Evaluation benchmarks organized by the APE hierarchy, aligned with corresponding PAR stages.}
\vspace{-0.1in}
\label{tab:eval-pyramid}
\end{table*}

Centered on these four questions, we organize MMR methods under a \textbf{Perception–Alignment–Reasoning} (PAR) framework, which decomposes MMR approaches into three interdependent stages: (1) \textbf{Perception},   extracting structured mathematical evidence from visual and textual modalities; (2) \textbf{Alignment},   mapping perceived facts to symbolic or executable representations; and (3) \textbf{Reasoning},   conducting interpretable and verifiable inference over the aligned representations (e.g., CoT, program execution, tool use).
To complement this process-centric perspective, we further introduce a companion evaluation hierarchy, the \textbf{Answer–Process–Executable} (APE) framework. APE assesses correctness at three levels, 
\emph{answer} (task accuracy), \emph{process} (faithfulness of intermediate reasoning steps), and \emph{executable} (verification via executable checks).
Together,  PAR and APE  provide a systematic lens for dissecting multimodal \emph{mathematical} reasoning
enabling both a comprehensive synthesis of prior work and a diagnostic understanding of where current MLLMs succeed or fail to reason faithfully.

The roadmap of this survey is shown in Figure~\ref{intro}. We begin by outlining the core challenges and preliminaries of MMR, including main task families and the structure of perception outputs. We then formalize the PAR pipeline and synthesize methods at each stage. For \emph{Perception}, we track the path from symbolic parsers to pipelines built on large multimodal models (Section \ref{sec:perception}). For \emph{Alignment}, we cover executable intermediates, symbolic and neural hybrids, cross-modal alignment frameworks, and pretraining and finetuning strategies (Section \ref{sec:q2-align}). For \emph{Reasoning}, we review deliberate chains, reinforcement learning, tool-augmented and executable reasoning, and process feedback and verification (Section \ref{sec:q3-reasoning}). Next, we map major benchmarks and datasets to APE levels and to PAR stages (Section \ref{sec:q5howtoevaluate}), and we provide consolidated tables for direct comparison and diagnostic analysis (Figure \ref{fig:taxonomy} and Tables~\ref{tab:eval-pyramid}-\ref{tab:task-par}).
We finally conclude the survey by outlining open challenges and future directions (Section \ref{sec:challenges}).

\section{Perception: What to Extract?}
\label{sec:perception}

In the PAR framework (overview shown in Figure \ref{fig:taxonomy}), perception addresses the first and central question, what to extract from multimodal inputs before alignment and reasoning can occur. Unlike generic vision tasks, mathematical perception must yield structured, computation relevant evidence rather than only objects or text. Given 
multimodal inputs, i.e., \(X \subseteq \{T,D,C,I\}\) a mixture of text $T$, diagram $D$, chart or table $C$,  and image $I$,   
the perception function $p: X \mapsto \mathcal{F}$ extracts a set of mathematical facts $\mathcal{F}$ spanning three levels: (i) low level primitives such as points, lines, axes, or objects, (ii) structural relations such as incidence, parallelism, axis series binding, or row and column layouts, and (iii) quantitative attributes such as lengths, angles, values, and units. Note that perception is essential; errors at this stage propagate downstream and can lead to misalignment or faulty reasoning.

To ground \emph{PAR} in concrete settings, we introduce  
three representative task families:  \emph{geometry problems, chart/table problems}, and \emph{visual math word problems}. These task families illustrate the kinds of evidence that must be extracted. We then summarize the task-oriented datasets through the lens of \emph{PAR} (detailed in Table~\ref{tab:task-par}), which provides the complete list of datasets for each task. Finally, we review the methodological evolution of perception, from symbolic parsers to neural encoders to LMM-based pipelines, and conclude with an outlook on open challenges and promising directions.

\begin{table*}[h]
\centering
\scriptsize
\tiny
\setlength{\tabcolsep}{4pt}
\resizebox{1.05\textwidth}{!}{%
\begin{tabular}{l l l l p{4.4cm}}
\toprule
\textbf{Dataset} & \textbf{Year (Venue)} & \textbf{PAR Stage} & \textbf{Size / Annotation} & \textbf{Key Contributions} \\
\midrule
\multicolumn{5}{c}{\textbf{Geometry Problem}} \\[-0.3ex]
\midrule
GEOS~\cite{seo-etal-2015-solving} & 2015 (EMNLP) & Perception + Alignment & 55 questions; text + diagram & early GPS baseline; text–diagram mapping \\
GEOS++~\cite{sachan-etal-2017-textbooks} & 2017 (EMNLP) & Alignment & 1{,}406 questions; partial logical forms & SAT-style benchmark with logical grounding \\
Geometry3K~\cite{lu2021intergpsinterpretablegeometryproblem} & 2021 (ACL) & Perception + Alignment & 3{,}002 questions; dense formal language & formal grounding linking text and diagrams \\
GeoQA~\cite{chen2022geoqageometricquestionanswering} & 2021 (ACL Findings) & Alignment + Reasoning & 5{,}010 questions; executable programs & program-supervised QA \\
GeoQA+~\cite{cao-xiao-2022-augmented} & 2022 (COLING) & Alignment + Reasoning & extended set with harder steps & challenging multi-step reasoning test \\
PGDP5K~\cite{hao2022pgdp5kdiagramparsingdataset} & 2022 (IJCAI) & Perception & 5,000 diagrams; primitive labels & dataset for geometric primitive parsing \\
PGPS9K~\cite{zhang2023multimodalneuralgeometricsolver} & 2023 (IJCAI) & All & 9,022 items; fine-grained diagram + program & interpretable diagram–program pairs \\
UniGeo~\cite{chen2022unigeounifyinggeometrylogical} & 2022 (EMNLP) & Alignment + Reasoning & 4,998 calc + 9,543 proofs & unified format covering calculation and proof \\
GeomVerse~\cite{kazemi2023geomversesystematicevaluationlarge} & 2024 (ICML Workshop)& Reasoning & procedurally generated problems & synthetic benchmark to test reasoning capacity \\
FormalGeo7K~\cite{zhang2024formalgeoextensibleformalizedframework} & 2024 (NeurIPS Workshop) & Alignment + Reasoning & $\sim$7,000 problems; diagram + formal solution & verifiable formal geometry tasks \\
Geo170K~\cite{gao2023gllavasolvinggeometricproblem} & 2025 (ICLR) & Perception + Alignment & $\sim$170,000 image–caption + QA pairs & large-scale geometry pretraining set \\
GeoGPT4V~\cite{cai2024geogpt4vgeometricmultimodallarge} & 2024 (EMNLP) & Perception + Alignment & 4,900 synthesized + 19,000 mixed pairs & LLM-generated geometry text–figure dataset \\
MATHGLANCE~\cite{sun2025mathglancemultimodallargelanguage} & 2025 (arXiv) & Perception & $\sim$1,200 diagrams/1,600 questions; perception tags & isolates perception-level evaluation \\
\midrule
\multicolumn{5}{c}{\textbf{Chart and Table Problems}} \\[-0.3ex]
\midrule
FigureQA~\cite{kahou2018figureqaannotatedfiguredataset} & 2018 (ICLR Workshop) & Perception & $\sim$100,000 charts; $\sim$1M QA & synthetic chart reasoning dataset \\
DVQA~\cite{kafle2018dvqa} & 2018 (CVPR) & Perception & $\sim$300,000 images; $>$3M QA & open vocabulary chart questions with metadata \\
PlotQA~\cite{methani2020plotqareasoningscientificplots} & 2020 (WACV) & Perception & 224,377 plots; $\sim$28.9M QA & real-valued numeric reasoning on scientific plots \\
ChartQA~\cite{masry2022chartqa} & 2022 (ACL Findings) & Perception + Alignment & 9,600 human + 23,100 generated QA & visual + logical chart QA \\
CharXiv~\cite{wang2024charxivchartinggapsrealistic} & 2025 (NeurIPS) & Perception & 2,323 curated charts & scientific chart understanding in real domain \\
ChartQAPro~\cite{masry-etal-2025-chartqapro} & 2025 (ACL) & Perception + Alignment & 1,341 charts with dashboards & more complex visualization types \\
ChartQA-X~\cite{hegde2025chartqaxgeneratingexplanationsvisual} & 2025 (arxiv) & Alignment & 30,299 charts with QA + rationale & supervision for explanation in charts \\
FinQA~\cite{chen2022finqadatasetnumericalreasoning} & 2021 (EMNLP) & Alignment + Reasoning & 8,281 cases with gold programs & hybrid table + text numerical reasoning \\
TAT-QA~\cite{zhu2021tatqaquestionansweringbenchmark} & 2021 (ACL) & Alignment + Reasoning & 16,552 QA in financial reports & table–text numerical reasoning benchmark \\
MultiHiertt~\cite{zhao-etal-2022-multihiertt} & 2022 (ACL)  & Alignment + Reasoning & 10,440 QAs in financial reports & hybrid table + text numerical reasoning \\
DocMath-Eval~\cite{zhao-etal-2024-docmath} & 2024 (ACL)  & Alignment + Reasoning & 4,000 QAs in financial reports; gold programs & hybrid table + text numerical reasoning \\
TabFact~\cite{chen2020tabfactlargescaledatasettablebased} & 2020 (ICLR) & Alignment & 118,000 statements; 16,000 tables & table entailment verification dataset \\
WikiTableQuestions~\cite{pasupat2015compositionalsemanticparsingsemistructured} & 2015 (ACL) & Alignment + Reasoning & 22,033 QA; 2,108 tables & compositional QA over web tables \\
WikiSQL~\cite{zhong2017seq2sqlgeneratingstructuredqueries} & 2017 (NeurIPS) & Alignment & 80,654 NL–SQL; 24,241 tables & executable SQL supervision benchmark \\
DUDE~\cite{vanlandeghem2023documentunderstandingdatasetevaluation} & 2023 (ICCV) & All & multi-page document datasets & document-level reasoning with table/figure content \\
\midrule
\multicolumn{5}{c}{\textbf{Visual Math Word Problems}} \\ [-0.3ex]
\midrule
IconQA~\cite{lu2021iconqa} & 2021 (NeurIPS) & Perception + Reasoning & 107,439 questions; multiple formats & large-scale multimodal math QA benchmark \\
Icon645~\cite{lu2021iconqa} & 2021 (NeurIPS) & Perception & 645,687 icons; 377 classes & icon pretraining resource \\
TABMWP~\cite{lu2023dynamicpromptlearningpolicy} & 2023 (ICLR) & Alignment + Reasoning & 38,431 problems; gold solutions / programs & table-based visual math word problems \\
CLEVR-Math~\cite{lindström2022clevrmathdatasetcompositionallanguage} & 2022 (NeSy) & Perception + Reasoning & synthetic image + text arithmetic & compositional arithmetic reasoning \\
MV-MATH~\cite{wang2025mv} & 2025 (CVPR) & All & 2,009 multi-image problems & cross-image dependency reasoning for K–12 \\
MathVista~\cite{lu2024mathvistaevaluatingmathematicalreasoning} & 2024 (ICLR) & All  & 6,000+ visual math problems; 28 merged sets & combining diagrams, charts, and images \\
MATH-V~\cite{wang2024measuring} & 2024 (NeurIPS) & All  & 3,040 curated visual problems & higher-difficulty multimodal reasoning benchmark \\
Math2Visual~\cite{wang-etal-2025-generating-pedagogically} & 2024 (ACL Findings) & Perception + Alignment & 12,000 generated visuals from math word text & benchmark for text-to-diagram generation in math \\
\bottomrule
\end{tabular}%
}\vspace{-0.1in}
\caption{Datasets grouped by task and annotated with the primary  PAR stage they support, plus year, venue, size, and key contributions.}
\label{tab:task-par}\vspace{-0.18in}
\end{table*}

\paragraph{Geometry Problems.}
Geometry problem solving requires models to jointly parse textual descriptions $T$ and diagrams $D$ to produce numerical values, symbolic relations, or complete proofs: $f:(T,D)\mapsto y$.
Perception in this task focuses on recognizing geometric primitives such as points, lines, and angles, understanding their spatial relations, and grounding textual references to diagrammatic structures before performing deductive reasoning.
Method development has progressed from symbolic theorem provers such as GEOS~\cite{seo-etal-2015-solving}, to neural vision–language models, and more recently to hybrid pipelines with executable programs such as E-GPS~\cite{wu2024gps} and Pi-GPS~\cite{zhao2025pi}, which enhance verifiability and explainability.
LMMs further introduce a new perception paradigm, enabling both improved geometric understanding, as seen in GeomVerse~\cite{kazemi2023geomversesystematicevaluationlarge}, and large-scale synthetic data generation, as demonstrated by G-LLaVA~\cite{gao2023gllavasolvinggeometricproblem} and GeoGPT4V~\cite{cai2024geogpt4vgeometricmultimodallarge}.
Recent work further explores \textit{diagram formalization} and formal-language pretraining to improve structural understanding and robustness under domain shift, such as DFE-GPS~\cite{xin2025generalizable} and GEOX~\cite{xia2024geox}.
Representative datasets include Geometry3K~\cite{lu-etal-2021-inter}, GeoQA and GeoQA+~\cite{chen2022geoqageometricquestionanswering,cao-xiao-2022-augmented}, PGDP5K~\cite{hao2022pgdp5kdiagramparsingdataset}, and PGPS9K~\cite{zhang2023multimodalneuralgeometricsolver}.

\paragraph{Chart and Table Problems.}
Chart and table problems assess the ability to interpret structured visual data $C$ in response to a natural language query $Q$, formalized as $f:(C,Q)\mapsto a$, where $a$ denotes the predicted answer.
Models must accurately perceive visual layouts such as axes, legends, rows, and columns, ground linguistic references to these visual elements, and perform numerical or logical reasoning based on the extracted structure.
Perception in this domain has evolved from explicit symbolic parsing~\cite{kafle2018dvqa,methani2020plotqa,masry2022chartqa} to neural vision–language models that jointly encode layout and text~\cite{lee2023pix2struct}, and more recently to LMM-based instruction-tuned frameworks~\cite{han2023chartllama, xia2024chartx} that integrate structural perception with executable reasoning. DePlot~\cite{liu2023deplotoneshotvisuallanguage} and LogicNLG~\cite{chen-etal-2020-logical} bridge perception and alignment through chart to table translation.
Key benchmarks include PlotQA~\cite{methani2020plotqareasoningscientificplots}, 
TATQA~\cite{zhu2021tatqaquestionansweringbenchmark}, FinQA~\cite{chen2022finqadatasetnumericalreasoning},
MultiHiertt~\cite{zhao-etal-2022-multihiertt}, 
ChartQA~\cite{masry2022chartqa}, and DocMath-Eval~\cite{zhao-etal-2024-docmath}.

\paragraph{Visual Math Word Problems.}
Visual Math Word Problems require solving natural-language math queries grounded in visual scenes: $f:(I,Q)\mapsto a$, where $Q$ denotes the natural-language question and $a$ denotes the predicted answer.
Typical skills include object counting, attribute reasoning, quantity comparison, and cross-image co-reference.
Methods have gradually shifted from symbolic perception and explicit object relation parsing like Patch-TRM ~\cite{lu2021iconqa}
to neural multimodal encoders that learn visual–textual correspondences~\cite{lu2021inter},
and more recently to LMMs capable of holistic scene understanding and chain-of-thought reasoning~\cite{cai2024geogpt4v}.
Representative datasets include IconQA~\cite{lu2021iconqa}, CLEVR-Math~\cite{lindstrom2022clevr}, TABMWP~\cite{lu2022dynamic}, RoMMath~\cite{zhao-etal-2025-multimodal}, and MV-MATH~\cite{wang2025mv}. 

\paragraph{Method Evolution and Outlook.}
Methods for mathematical perception have progressed from symbolic parsers and handcrafted rules to neural encoders that couple visual grounding with textual understanding, and now to LMMs unified through pretraining and instruction tuning.  
Despite their generality, LMMs often struggle with fine-grained perception, such as misreading geometric elements or chart layouts.
Future work should focus on precise structure perception, executable supervision, and combining neural and symbolic reasoning for reliable results.

\section{Alignment: How to Represent \& Align?}
\label{sec:q2-align} 

Alignment bridges perception and reasoning. It defines how perceived visual facts are structured and mapped to symbolic or linguistic forms so that downstream reasoning becomes interpretable and verifiable. In mathematical contexts, alignment connects visual entities such as geometric primitives, chart axes, and table layouts with textual predicates or executable intermediates like geometry description languages, constraint sets, proof sketches, chart or table operators, SQL queries, and program-of-thought traces. The key challenge is to represent and align multimodal information while preserving symbolic fidelity and remaining robust to visual noise and domain variation. This section reviews alignment techniques from four complementary perspectives: (1) \emph{executable intermediates} that formalize visual content into checkable programs, (2) \emph{symbolic–neural hybrids} that couple neural perception with symbolic reasoning engines, (3) \emph{cross-modal frameworks} that stabilize vision–language coupling, and (4) \emph{pre-training and fine-tuning strategies} that provide large-scale priors and task-specific supervision. 

\subsection{Executable Intermediates}
\label{executableintermediates}
A key direction is converting visual content into formal, checkable intermediates that support symbolic reasoning. Inter-GPS~\cite{lu2021inter} annotate geometry problems with domain-specific languages to enable interpretable execution. E-GPS~\cite{wu2024gps} integrates a symbolic solver with a diagram parser for verifiable step-by-step solutions. Pi-GPS~\cite{zhao2025pi} introduces a multimodal rectifier to disambiguate diagrams before theorem-driven solving. R1-OneVision~\cite{yang2025r1} scales this idea by transforming diagrams into textual formalizations for large-scale consistency training.
Beyond geometry, chart and table reasoning convert visual marks into code- or SQL-like operators to ensure numeric correctness by design. Executable intermediates thus anchor alignment and make reasoning verifiable.

\subsection{Symbolic--Neural Hybrids}
\label{symbolicneuralhybrids}
Hybrid pipelines combine symbolic rigor with neural flexibility. For example, GeoGen~\cite{pan2025enhancing} aligns diagrams with executable programs under symbolic supervision. MathCoder-VL~\cite{wang2025mathcoder} uses code-based cross-modal supervision to reinforce visual and text alignment and program-level faithfulness. AlphaGeometry~\cite{trinh2024solving} integrates theorem libraries with neural search to handle complex geometric deductions.
By injecting formal structure while retaining perceptual capacity, these hybrids enhance interpretability, transferability, and reasoning stability.

\subsection{Cross-modal Alignment Frameworks}
\label{crossmodalalignmentframworks}
General frameworks provide reusable backbones for stable vision–language coupling. BLIP-2~\cite{li2023blip} links vision encoders to LLMs and serves as a base for math-specific extensions. LLaVA~\cite{liu2023visual} introduces instruction-following alignment for visual inputs. Math-PUMA~\cite{zhuang2024mathpumaprogressiveupwardmultimodal} applies progressive staged alignment for long-chain stability, while VCAR~\cite{jia2024describe} follows a “describe-then-reason” curriculum.
For long-horizon reasoning, TVC~\cite{sun2025mitigating} maintains persistent visual conditioning, and VIC~\cite{zheng2024thinking} composes textual plans with late fusion to avoid drift. Curriculum- and conditioning-based designs help reduce cumulative errors and stabilize multi-step reasoning.

\subsection{Pre-training and Fine-tuning as Enablers}
\label{pretrainandfinetuneingasenablers}
Large-scale pre-training provides broad coverage and alignment priors. Geo170K~\cite{gao2023g}, SynthGeo228K~\cite{zhang2025diagram}, TrustGeoGen~\cite{fu2025trustgeogen} and GeoGPT-4V~\cite{cai2024geogpt4v} expand diagram–text coupling at scale. Math-LLaVA~\cite{shi2024math} and MAVIS~\cite{zhang2024mavis} extend instruction-tuned data with visual reasoning. MultiMath-300K~\cite{peng2024multimath} contributes multimodal K–12 problems with stepwise annotations.Beyond these, MAmmoTH-VL~\cite{guo2024mammoth} scales to 12M instruction pairs for multimodal pre-training, while ~\cite{fu2025trustgeogen} generates verified geometric data for reliable training.
Symbolic resources like AlphaGeometry~\cite{trinh2024solving} and auto-diagram construction~\cite{krueger2021automatically} further enhance formal priors.
Objective design mixes grounding with process supervision—Masked Thought~\cite{chen2024masked} learns from partial steps, LogicSolver~\cite{yang2022logicsolver} integrates logical constraints, and MathGenie~\cite{lu2024mathgenie} generates synthetic CoT data.

Fine-tuning specializes alignment toward executable reasoning. MMathCoT-1M and DualMath-1.1M~\cite{shi2024math,zhang2024mavis} link QA with dual-view trajectories, while MathV360K~\cite{shi2024math} and MAVIS~\cite{zhang2024mavis} provide diagram-based instruction data. Datasets such as Geometry3K~\cite{lu2021inter}, GeoQA~\cite{chen2021geoqa}, and E-GPS~\cite{wu2024gps} enable symbolic supervision and program-level verifiability.
Curricular designs like VCAR~\cite{jia2024describe}, Math-PUMA~\cite{zhuang2024mathpumaprogressiveupwardmultimodal}, and AtomThink~\cite{xiang2024atomthink} progressively refine perception and reasoning, making alignment robust and transferable.

\paragraph{Outlook and Comparison.} Executable intermediates  ensure verifiability but are brittle under domain shifts. Symbolic–neural hybrids improve robustness yet add complexity. Cross-modal frameworks scale well but risk inconsistencies without explicit execution. Pre-training and fine-tuning bring generality but depend on data fidelity. In practice, combining executable precision, hybrid robustness, curriculum stability, and large-scale priors can perhaps achieve the best balance between reliability and generalization.

\section{How to perform Reasoning?}
\label{sec:q3-reasoning} 

After perception and alignment produce structured representations, the final stage concerns how models perform reliable inference.
Reasoning in multimodal mathematical tasks involves executing stable and verifiable computation from structured inputs. Four paradigms dominate: (1) \emph{Deliberate chain (e.g., CoT) methods}, which externalize intermediate steps to expose and guide reasoning; (2) \emph{Reinforcement learning methods}, which optimize long-horizon decision sequences via reward-guided search; (3)  \emph{Tool-augmented reasoning}, which employs external solvers or code execution to enforce formal correctness; and (4) \emph{Process feedback and verification}, which introduces critics or verifiers to assess intermediate steps (e.g., executable checks, self-consistency), improving validity and interpretability.
These approaches collectively enhance robustness and faithfulness across long reasoning chains.
Beyond these main paradigms, \emph{Error Detection and Correction} (to flag and repair faulty traces) and \emph{Mathematical Problem Generation}  (to synthesize diverse, curriculum-aligned instances) play supportive roles that strengthen process supervision and dataset curation. Due to space limits, we defer discussion of these topics to Appendix \ref{SupervisionReasoning}.

\subsection{Deliberate Chains (e.g., Chain-of-Thought)}
\label{deliberatechains}
In-Context Learning (ICL) with multimodal chain-of-thought (CoT) prompts models to externalize intermediate steps. LLaVA-CoT \cite{xu2024llava} shows that structured prompts can elicit more reliable reasoning paths. TVC \cite{sun2025mitigating} injects persistent visual conditioning at every step to mitigate forgetting. VIC \cite{zheng2024thinking} composes plans in text first and fuses vision later to reduce cross-modal drift. I2L \cite{wang2024all} embeds exemplars directly on the visual canvas to strengthen grounding. AtomThink \cite{xiang2024atomthink} decomposes reasoning into atomic steps, improving compositionality and enabling fine-grained supervision. Although these methods are lightweight and effective, they can still drift away from the underlying evidence without stronger grounding or verification mechanisms.

Beyond linear chains, Tree of Thoughts (ToT) \cite{yao2023tree} generalizes CoT by exploring and self-evaluating multiple branches of intermediate thoughts, and Graph of Thoughts (GoT) \cite{besta2024graph} further models non‑linear dependencies among partial solutions. For multimodal settings, AGoT \cite{yang2024soft} adapts GoT to multi‑modal representation learning via an aggregation graph that soft‑prompts and routes reasoning across aspects. For multimodal mathematical reasoning specifically, VisuoThink \cite{wang2025visuothink} performs multimodal tree search with interleaved vision–text steps, and VReST \cite{zhang2025vrest} combines Monte Carlo Tree Search with a self‑reward signal to deepen exploration and reports state‑of‑the‑art results on several multimodal math benchmarks. Together, these ToT/GoT‑style methods complement CoT by enabling branching, backtracking, and structured selection over intermediate solutions, which is valuable for long‑horizon visual–symbolic math problems.

\subsection{RL-based Reasoning}
\label{rlbasedreasoning}
Reinforcement learning (RL) approaches treat reasoning as a sequential decision process and optimize for long-horizon stability.

\paragraph{Reward Mechanism Design.}  
R1-VL~\cite{zhang2025r1} introduces step-wise accuracy and validity rewards to encourage high-quality transitions.
VisualPRM~\cite{wang2025visualprm} learns Process Reward Models (PRMs) from large-scale multimodal supervision to provide dense step-level feedback.
MM-PRM~\cite{du2025mm} combines PRM supervision with Monte Carlo Tree Search (MCTS) for comprehensive evaluation.
MM-Eureka~\cite{meng2025mm} explores rule-based RL to capture “visual aha” moments with minimal human annotation.

\paragraph{Search and Decision Algorithms.}  
DeepSeek-R1~\cite{guo2025deepseek} applies Group Relative Policy Optimization (GRPO) to jointly optimize reasoning and search, and Vision-R1~\cite{huang2025vision} extends this to multimodal settings.
Mulberry~\cite{yao2412mulberry} integrates MCTS with reflective reasoning for iterative correction, while Skywork R1V2~\cite{chris2504skywork} combines Maximum a Posteriori Policy Optimization (MPO) and GRPO to balance detail and generalization.
VL-Rethinker~\cite{wang2025vl} uses selective sample replay to mitigate vanishing advantages.
FAST~\cite{xiao2025fast} adapts inference depth to question complexity, and Think-or-Not?~\cite{wang2025think} learns when to engage in deep reasoning.
VLAA-Thinking~\cite{chen2025sft} studies reflection-aware optimization and contrasts RL with Supervised Fine-Tuning (SFT).
VLM-R$^3$~\cite{jiang2025vlm} proposes a three-stage pipeline of region recognition, reasoning, and refinement, while MAYE~\cite{ma2025rethinking} and SoTA-with-Less~\cite{wang2025sota} focus on sample efficiency via MCTS-guided data selection. Beyond multimodal reasoning, AlphaProof~\cite{deepmind2024alphaproof} extends reinforcement learning to formal theorem proving via self-play and symbolic verification in Lean, achieving silver-medal performance on IMO problems. It exemplifies how RL can support verifiable and executable mathematical reasoning.

\subsection{Tool-Augmented Reasoning}
\label{toolaugmentedreasoning}
Tool-augmented methods delegate parts of reasoning to external symbolic systems or APIs to enhance modularity and correctness.
Toolformer~\cite{schick2023toolformer} demonstrates how LLMs can invoke external tools for symbolic computation and retrieval, while ToRA~\cite{gou2023tora} organizes iterative loops of reasoning, tool calls, and result integration.
COPRA~\cite{thakur2023context} composes multiple external capabilities adaptively, and MM-REACT~\cite{yang2023mm} coordinates visual and textual tools for multimodal reasoning.
For geometry, Visual Sketchpad~\cite{hu2024visual} provides an interactive canvas that enables models to construct and reason visually, and Pi-GPS~\cite{zhao2025pi} integrates parsers, verifiers, and symbolic solvers to produce provable results.
Chameleon~\cite{lu2023chameleon} illustrates dynamic multi-tool composition, while MathCoder-VL~\cite{wang2025mathcoder} uses code supervision to align diagrams with programs, making reasoning directly executable.
Together, these systems show how tool integration supports structured, verifiable, and interpretable reasoning.

\subsection{Process Feedback and Verification}
\label{processfeedbackandverification}
VisualPRM \cite{wang2025visualprm} provides process-level rewards that encourage valid steps and penalize errors. MM-PRM \cite{du2025mm} integrates PRM scoring with search, creating a generate–judge–revise loop for stable chains. Proof and program verifiers check intermediate Domain-Specific Language, code, or proof sketches, ensuring results are executable. At the representation level, TVC \cite{sun2025mitigating} maintains visual conditioning during reasoning, while VIC \cite{zheng2024thinking} reduces bias by text-first planning and late fusion. These approaches connect training with evaluation, ensuring that models are judged not only by answers but also by the correctness of their processes.

\paragraph{Outlook and Comparison.} 
Different reasoning paradigms show complementary strengths. Deliberate chains are lightweight but risk drifting from visual evidence. Reinforcement learning stabilizes long reasoning yet demands costly rewards. Tool-augmented methods add modularity and verifiability but rely on stable interfaces. Process feedback improves auditability but needs dense supervision. Overall, hybrid systems that combine explicit reasoning chains, selective reinforcement learning, executable intermediate representations, and verification mechanisms appear especially promising for robust and interpretable multimodal reasoning.

\section{How to Evaluate? }
\label{sec:q5howtoevaluate}
To distinguish genuine mathematical reasoning from shortcut use, evaluation must span the full \textbf{PAR} pipeline and follow our \textbf{Answer–Process–Executable (APE)} hierarchy. 
\textbf{Answer}:  Final-task metrics (e.g., accuracy) that are easy to report but can conflate perception errors (e.g., misread diagrams) and alignment errors (e.g., incorrect bindings) with reasoning mistakes.
\textbf{Process}: Step-level checks that test whether intermediate reasoning is valid and \emph{visually grounded} (i.e., consistent with extracted primitives and relations).
\textbf{Executable}:  Faithfulness via execution or proof checking (e.g., running code, verifying constraints/derivations) to directly assess alignment and reasoning correctness.
We summarize how existing benchmarks map to the \textbf{APE} dimensions in Table~\ref{tab:eval-pyramid}.
The table also covers \textbf{Comprehensive} benchmarks (see Appendix~\ref{comprehensivebenchmarks}) that combine diverse modalities, tasks, and difficulty levels to assess overall reasoning ability. 
Other benchmarks, including robustness (e.g., probing sensitivity to visual perturbations) and domain-specific sets (e.g., remote sensing), are discussed in Appendix~\ref{app:robust}.

\subsection{Answer-level Evaluation}
\label{answerlevelevaluation}
Answer-level benchmarks judge the final answer with exact match or numeric tolerance. ChartQA \cite{masry2022chartqa} evaluates reasoning over diverse real-world charts; PlotQA \cite{methani2020plotqareasoningscientificplots} stresses open-vocabulary and real-valued answers on scientific plots; FigureQA \cite{kahou2017figureqa} provides large-scale synthetic charts for controlled visual reasoning. IconQA \cite{lu2021iconqa} assesses icon-like visual math with multiple formats and cognitive skills. CLEVR-Math \cite{lindstrom2022clevr} probes compositional arithmetic in synthetic scenes. Hybrid table–text datasets such as FinQA \cite{chen2022finqadatasetnumericalreasoning} and TAT-QA \cite{zhu2021tatqaquestionansweringbenchmark} evaluate numerical reasoning over structured evidence. Answer-level evaluation is scalable and task-agnostic but cannot separate lucky guesses from correct reasoning, nor does it reveal where the Perception, Alignment and Reasoning pipeline failed.

\subsection{Process-level Evaluation}
\label{processlevelevaluation}
Process-level benchmarks attach or elicit intermediate steps and score their validity, shifting the focus from answers to how solutions are produced. MM-MATH \cite{sun2024mm} provides step types and error annotations on middle-school problems with visual contexts. MPBench \cite{xu2025mpbench} evaluates step-level judges and finds that many general multimodal models struggle with systematic error identification. ErrorRadar \cite{yan2024errorradar} contributes fine-grained error taxonomies and labels for diagnostic analysis, and Sherlock \cite{ding2025sherlock} extends multimodal process diagnosis with detailed failure categories. We-Math~\cite{qiao2024we} emphasizes principle-centered process evaluation beyond end-to-end scores, MathVerse~\cite{zhang2024mathverse} perturbs diagrams to test visual understanding beyond text priors, CHAMP~\cite{mao2024champ} annotates concepts and hints and reports cases where models reach correct answers with wrong steps, and PolyMATH~\cite{gupta2024polymath} covers diverse cognitive categories including spatial and pattern reasoning. These resources enable audits of faithfulness and robustness while exposing where Perception or Alignment drifts translate into faulty Reasoning steps.

\subsection{Executable-level Evaluation}
\label{executablelevelevaluation}
Executable-level benchmarks require programs, proofs, or constraints that can be run or verified, directly testing symbolic Alignment and the faithfulness of Reasoning. GeoQA+ \cite{cao2022augmented} annotates step-by-step programs for geometry and validates them by execution. FormalGeo \cite{zhang2024formalgeoextensibleformalizedframework} offers Olympiad-level geometry with formal statements, theorem sequences, and verifiable proofs. Inter-GPS and E-GPS \cite{lu2021inter,wu2024gps} provide formal languages and solver-backed pipelines, and Pi-GPS \cite{zhao2025pi} adds an LMM rectifier with a theorem-driven solver to produce provable chains. Executable metrics give clear pass or fail results that help identify alignment or reasoning errors, but they depend on reliable parsers and checkers.

\section{Challenges and Future Directions}
\label{sec:challenges}

MMR has advanced rapidly, yet key challenges remain.  
Following the PAR framework, we summarize major limitations and future directions.

\paragraph{Perception.}  
Current MLLMs show only a shallow understanding of visual information and often fail under layout or style changes \cite{liu2025rolevisualmodalitymultimodal,liu2025role}.  
Structured diagram parsing that captures primitives, topology, and layout improves robustness \cite{wu2024gps}.  
A promising direction is to pair structured perception with formal interfaces such as code, proof sketches, or SQL, enabling visual evidence to be verified through execution \cite{zhao2025pi,lu2021inter}.

\paragraph{Alignment.}  
Fragmented \textit{domain-specific languages (DSLs)} and inconsistent unit conventions cause misalignment and limit transfer.  
Future work should design unified, type-aware DSLs with explicit unit handling, constraint checking, and program verification \cite{pan2025enhancing} to standardize visual–symbolic mappings.

\paragraph{Reasoning.}  
Long reasoning chains tend to drift from visual evidence.  
 RL improves stability but is expensive and sensitive to reward design.  
Lightweight reward models, adaptive inference depth, and hybrid pipelines that delegate symbolic steps to external verifiers can reduce cost while maintaining robustness \cite{guo2025deepseek,huang2025vision,wang2025vl,wang2025visualprm}. This reflects a broader trade-off between stability and cost reinforcement learning enhances consistency but introduces heavy computational demands, motivating lightweight process rewards and symbolic verification for practical scalability.
However, benchmark-based evaluation remains limited: models may overfit to specific datasets or annotation styles rather than acquiring transferable reasoning skills.  
True reasoning should extend beyond curated benchmarks to unseen problems and open-ended contexts \cite{liang2024scemqa,cherian2024evaluating}. 

\paragraph{Future Opportunities.}  
Applications such as intelligent tutoring, automated grading, and theorem explanation can enhance education through process-aware feedback \cite{zhou2024mathscape,ku2025theoremexplainagent,du2025mm}.  
Accessibility tools like MathCAT and MathVision translate visual math into speech or braille with executable checks for accuracy \cite{soiffer2024mathcat,awais2024mathvision}.  
Professional systems for AR, VR, and engineering can integrate sketchpads, solvers, and code interfaces for verifiable design \cite{hu2024visual}.  
Advancing these directions while addressing PAR-level challenges will lead to more reliable and interpretable multimodal reasoning systems.
Detailed discussions on challenges and future opportunities are provided in Appendix~\ref{Challenges:Future}.

\section{Conclusion} 
This paper presents a process-centered framework of MMR built on the Perception–Alignment–Reasoning (PAR) pipeline and the Answer–Process–Executable (APE) hierarchy. By organizing progress across geometry, chart and table reasoning, and visual math word problems, we show how structured perception, symbolic alignment, and verifiable reasoning jointly enable reliable multimodal intelligence. The PAR and APE frameworks offer a unified lens for understanding methods, benchmarks, and open issues, emphasizing structure-aware perception, executable intermediates, and process-level evaluation.


\bibliographystyle{unsrtnat}
\bibliography{custom}

\newpage

\appendix
\section*{Appendix}

\begingroup
\setcounter{table}{0}
\renewcommand{\thetable}{A\arabic{table}} 
\endgroup



\begin{table*}[h]
\centering
\scriptsize
\setlength{\tabcolsep}{4pt}
\resizebox{\textwidth}{!}{
\begin{tabular}{l l l c p{6cm}}
\toprule
\textbf{Survey} & \textbf{Venue \& Year} & \textbf{Scope / Focus} & \textbf{Models} & \textbf{Focus} \\
\midrule
\citet{lu-etal-2023-survey} & ACL’23 & DL4Math & Deep Learning & Pre-LLM; model architectures and datasets; 
\\
\citet{10.1145/3729218} & ACM Computing’25 & FM4Reason & MLLM & Broad reasoning (limited math/symbolic depth)\\
\citet{ahn2024large} & EACL Workshop’24 & LLM4Math & LLM & Text-centered (non-MMR) \\
\citet{yan2024survey} & ACL Findings’25 & MLLM4Math & MLLM & Benchmark- and Model-centric taxonomy  \\ 
\citet{li2025perception} & arXiv’25 & LMRM  & LLM/MLLM & Roadmap- and Stage-centric analysis\\
\textbf{Ours} & - & \textbf{MLLM4Math} & \textbf{MLLM} & \textbf{First unified process-level framework revealing internal mechanisms of multimodal mathematical reasoning} \\
\bottomrule
\end{tabular}}
\caption{Comparisons between representative surveys and ours. “Models” column indicates model scope discussed in each survey (e.g., deep learning models, LLM, MLLM).}
\label{tab:Comparisons}
\end{table*}

\section{Related Surveys}
\label{Related Surveys}
As shown in Table \ref{tab:Comparisons}, we summary recent related surveys. Recent surveys have examined mathematical reasoning and multimodal intelligence from complementary perspectives but differ in focus and depth. 
\citet{lu-etal-2023-survey} reviewed deep learning for mathematical reasoning, summarizing architectures and datasets in the pre-LLM era but without multimodal or process-level analysis. 
\citet{10.1145/3729218} broadly discussed reasoning with foundation models across commonsense, logical, and mathematical domains, yet its treatment of symbolic and multimodal reasoning remains superficial. 
\citet{ahn2024large} analyzed LLM-based mathematical reasoning through four dimensions: tasks, methods, factors, and challenges, offering a structured text-centered view but overlooking visual grounding and reasoning processes. 
\citet{yan2024survey} extended this to the multimodal large language model (MLLM) era, organizing research by benchmarks, methodologies, and challenges, and introducing model roles as Reasoner, Enhancer, and Planner. 
However, its emphasis lies on ecosystem taxonomy rather than the internal mechanism connecting perception and symbolic alignment. 
\citet{li2025perception} surveyed large multimodal reasoning models (LMRMs) and proposed a developmental roadmap from modular perception to agentic reasoning, integrating reinforcement learning and multimodal chain-of-thought. 
Although comprehensive in scope, it treats mathematics as one application and lacks formal analysis of symbolic-numeric grounding or verifiability. 

In contrast, our survey focuses specifically on multimodal mathematical reasoning (MMR), abstracting the workflow into the Perception–Alignment–Reasoning (PAR) framework and the Answer–Process–Executable (APE) evaluation hierarchy. 
Together, PAR and APE provide a unified lens for understanding how multimodal evidence is perceived, aligned, and executed in verifiable reasoning. 
This framework bridges the symbolic–neural perspective of early deep learning, the text-based view of LLM reasoning, and the model-centric paradigm of MLLMs, offering the first process-level synthesis of multimodal mathematical reasoning.

Overall, previous surveys remain largely descriptive and domain-specific, while ours advances toward a process-level, verifiable, and multimodal understanding of mathematical reasoning that integrates perception, alignment, and reasoning within a coherent analytical framework.

\section{Reasoning Pipeline: Perception, Alignment and Reasoning}
\label{sec:par}

We abstract multimodal math reasoning into three stages. This view clarifies where systems fail and how to design robust solutions.

\paragraph{Perception.} The goal is to recover computationally relevant visual facts. In geometry this means primitives and topology such as points, lines, angles, incidence, and equality. In charts and tables this means axes, legends, marks, tick reading, cell structure, and semantic units. Robust OCR and layout also matter in document settings. Errors at this stage, such as missed intersections or misread scales, often cascade.

\paragraph{Alignment.} The next step is to bind visual facts to textual predicates or to an intermediate representation that can be executed. Examples include a geometry description language, a set of constraints, a proof language, a sequence of operators for charts and tables, a SQL query, or a program of thought trace. Alignment benefits from explicit anchors and structural losses, from code or program supervision, and from formal interfaces. To reduce cross modal drift during long chains of thought, recent strategies first compose reasoning in text and then consult visual evidence, or maintain visual conditioning throughout the chain.

\paragraph{Reasoning.} The final step executes arithmetic, logic, theorem sequences, or programs, often with tool use such as calculators, symbolic solvers, or retrieval. Process level critics and rewards and search methods such as best of N or tree search help maintain validity over long chains. Retaining visual evidence and controlling bias are important for stability. In geometry, staged planning with verifier backed steps is especially effective.

This decomposition also guides evaluation. Some benchmarks focus on perception and alignment such as chart reading or primitive extraction. Others emphasize executable and checkable inference such as geometric proofs or program execution.

\section{Supervision and Data for Reasoning}
\label{SupervisionReasoning}
\subsection{Error Detection and Correction}
In multimodal mathematical reasoning, inference often involves long chains of cross-modal steps, which requires not only evaluating the final answer but also supervising and revising intermediate reasoning states. VisualPRM \cite{wang2025visualprm} provides process-level rewards with dense supervision, encouraging valid reasoning transitions and penalizing deviations. MM-PRM \cite{du2025mm} integrates PRM scoring with Monte Carlo Tree Search to form a generate–judge–revise loop that stabilizes long reasoning chains. Mathador-LM \cite{kurtic2024mathador} instantiates critique-driven revision for math solutions, promoting self-correction during inference. VATE \cite{xu2025ai} targets classroom drafts with interactive feedback loops aligned with human pedagogy. Sherlock \cite{ding2025sherlock} contributes fine-grained error taxonomies for process diagnosis, and ErrorRadar \cite{yan2024errorradar} provides labeled categories to localize typical failure modes. MM-MATH \cite{sun2024mm} supplies large-scale step and error annotations, while MPBench \cite{xu2025mpbench} shows that general-purpose multimodal models still struggle with systematic error identification. Together, these systems and resources operationalize step-level judging and correction, so models are evaluated and improved by how they reason, not just by final answers. 

\subsection{Mathematical Problem Generation}
In multimodal mathematical reasoning, generating high-quality problems is essential for driving model training and evaluation, especially by supplying process- and execution-level testbeds for perception, alignment, and reasoning. GeoGen \cite{pan2025enhancing} follows a generate–solve–verify loop coupling symbolic solvers with natural-language verbalization to guarantee checkable solutions. GeoGPT-4V \cite{cai2024geogpt4v} co-generates aligned text–figure pairs with a strong multimodal model to broaden geometric coverage. Math-LLaVA with MathV360K \cite{shi2024math} extends instruction-style data toward visual math, and MAVIS \cite{zhang2024mavis} provides an automatic data engine with chain-of-thought supervision for large-scale synthesis. MultiMath-300K \cite{peng2024multimath} curates K–12 multimodal problems with captions and stepwise solutions for process-aware training. AtomThink \cite{xiang2024atomthink} offers long atomic chains of thought to supervise compositional reasoning, while MathCoder-VL \cite{wang2025mathcoder} uses code as supervision to align diagrams with executable programs for verifiable generation. These generation pipelines and corpora supply controllable, diverse, and executable data that strengthen perception and alignment while furnishing robust evaluation environments.

\section{Robustness and Domain-specific Benchmarks}
\label{app:robust}
Robustness benchmarks probe sensitivity to visual perturbations, multi-image dependencies, and domain shifts beyond standard evaluation. VCBench \cite{li2024vcbench} focuses on explicit multi-image reasoning dependencies. DynaMath \cite{zou2024dynamath} applies dynamic perturbations to test shortcut reliance. HC-M3D \cite{liu2025role} constructs near-duplicate images that flip correct answers to measure vision dependence. SMART-840 \cite{cherian2024evaluating} collects K–12 visuo-linguistic problems to assess fundamental multimodal skills under varied conditions. Domain specific sets such as GeoMath \cite{zhougeomath} target remote-sensing imagery and subject-specific math tasks, while MV-MATH \cite{wang2025mv} extends multi-image reasoning to K–12 contexts. Together these datasets assess model stability, generalization, and cross-domain transfer for multimodal mathematical reasoning.

\section{Comprehensive Benchmarks}
\label{comprehensivebenchmarks}
Comprehensive suites mix modalities, tasks, and difficulties to profile broad capabilities. 
MathVista~\cite{lu2024mathvistaevaluatingmathematicalreasoning} aggregates problems from many sources spanning natural images, diagrams, and charts. 
MATH-V~\cite{wang2024measuring} emphasizes difficulty calibration and curated coverage across subjects. 
SceMQA~\cite{liang2024scemqa} introduces a scientific multimodal QA benchmark at the college entrance level including Mathematics and other core subjects to evaluate reasoning across disciplines. 
MM-K12~\cite{du2025mm} targets K–12 education scenarios with verifiable multimodal problems, bridging visual understanding and curriculum-level reasoning. 
OlympiadBench~\cite{cherian2024evaluating} reports expert-level annotations enabling stepwise evaluation on competition-grade math and physics, while the Children's Olympiads benchmark~\cite{he2024olympiadbench} evaluates reasoning on competition problems designed for younger students. 
MathScape~\cite{liang2024mathscape} focuses on photo-based scenarios with hierarchical categories and multi-dimensional evaluation. 
CMM-Math~\cite{liu2024cmm} extends these benchmarks to the Chinese language setting, highlighting multilingual reasoning capabilities. 
These suites provide breadth and coverage but often entangle perception, alignment, and reasoning in a single score.

\section{Challenges and Future Directions}
\label{Challenges:Future}
\subsection{Challenges}
\label{Challenges}
\paragraph{Evaluation Challenges.}
While the proposed Answer–Process–Executable (APE) evaluation level provides a structured lens for assessing reasoning fidelity, the executable-level evaluation remains challenging to scale.  
Current executable benchmarks such as GeoQA+~\cite{cao2022augmented}, FormalGeo~\cite{zhang2024formalgeoextensibleformalizedframework}, and Pi-GPS~\cite{zhao2025pi} depend on domain-specific languages, symbolic solvers, or theorem checkers that are largely confined to geometry or table reasoning tasks.  
Generalizing these pipelines to broader multimodal reasoning such as chart interpretation, visual word problems, or scientific document understanding requires unified annotation protocols and lightweight verification schemes.  
Moreover, executable evaluation often introduces heavy computational costs and relies on manually curated programs or proofs, limiting its practicality for large-scale MLLM assessment.  
Future work may explore scalable formal interfaces and semi-automated checkers that balance verifiability, coverage, and efficiency within the APE framework.

\paragraph{Cross-cutting Challenges.}  
Data contamination, limited reproducibility, safety, and interpretability remain persistent issues.  
Leakage audits, standardized reporting, and verifier-backed pipelines can improve reliability.  
Executable intermediates, process judges, and proof or code verification support interpretability and trustworthy reasoning \cite{hu2024visual}.

\subsection{Future Opportunities}
\label{applications}
Multimodal mathematical reasoning enables diverse downstream applications that benefit from the model's ability to process and integrate visual and symbolic modalities. We categorize representative applications into three core areas:

\noindent
\textit{\underline{1. Education and Learning.}} \
Education applications benefit greatly from multimodal reasoning. For example, in STEM learning, tools like TheoremExplainAgent \cite{ku2025theoremexplainagent} visually and symbolically guide students through theorems and problem-solving processes. Intelligent tutoring systems \cite{du2025mm} dynamically adapt based on student input, providing feedback by analyzing both diagrams and text. Automated grading systems \cite{zhou2024mathscape} can assess multi-step, visual-rich student solutions, improving evaluation accuracy and scalability.

\noindent
\textit{\underline{2. Accessibility and Inclusivity.}} \
For learners with disabilities, multimodal reasoning systems enable accessible content delivery. MathCAT \cite{soiffer2024mathcat} and Mathvision \cite{awais2024mathvision} translate visual math into speech and braille, facilitating interaction with geometry or charts. These systems also support alternative input/output modalities (e.g., voice, haptics), ensuring inclusive engagement with mathematical content.

\noindent
\textit{\underline{3. Professional and Interactive Systems.}} \
In real-world problem-solving tasks—such as data analysis, architecture, or engineering—professionals must reason over both visual schematics and textual instructions. Multimodal reasoning aids this integration. In parallel, interactive interfaces in AR/VR environments \cite{hu2024visual} allow users to engage with math through gestures, voice commands, or immersive visual aids. These interfaces, when empowered by multimodal reasoning, enhance spatial understanding and application-specific interaction.

\section{Interplay between Datasets, Models, and Evaluation.}
\label{app:interplay}

The PAR and APE frameworks imply that datasets, model architectures, and evaluation protocols are not independent choices. What a benchmark annotates, and at which APE level, largely determines which stage of the Perception–Alignment–Reasoning pipeline is stressed; in turn, emerging modeling paradigms reveal gaps in existing benchmarks. Answer-only suites such as MathVista~\cite{lu2023mathvista} and MATH‑V~\cite{wang2024measuring} mainly report final accuracy on static diagrams, charts, and scenes. Under this setting, models often combine one-shot perception with generic CoT or program-of-thought decoding, and answer-level RL can already improve aggregate scores, but perception, alignment, and reasoning failures are entangled and shortcut strategies remain hard to diagnose.

Process-oriented and robustness benchmarks make these interactions more explicit. We‑Math~\cite{qiao2024we} decomposes problems into concept-level sub-questions and reports IK/IG/CM/RM metrics, directly probing where knowledge and generalization fail along the reasoning chain. MathVerse~\cite{zhang2024mathverse} and related variants perturb diagrams or isolate text-only views to test whether models truly rely on visual evidence rather than textual priors. FlowVerse~\cite{chen2025mathflowenhancingperceptualflow} further factorizes problem information into DI/EI/RP/OQ versions and introduces FlowVerse‑CoT‑E, tying evaluation to step-level reasoning grounded in perceptual information. Dynamic benchmarks such as DynaMath~\cite{zou2024dynamath} complement this by generating multiple visual and textual variants per seed question and comparing average- vs worst-case accuracy, emphasizing robustness under benign perturbations rather than single-shot success. Together with process-annotated corpora such as MultiMath‑300K~\cite{peng2024multimath}, these resources naturally favor step-aware supervision (e.g., PRMs, RL with process or outcome rewards, search-based refinement) and make Perception, Alignment, and Reasoning errors more observable.
Executable- or program-level supervision further pushes models toward modular pipelines. Geometry datasets with DSLs, proofs, and solver-backed checks support systems that first convert diagrams into executable representations before reasoning. MathFlow and FlowVerse exemplify this trend in visual math: FlowVerse exposes which parts of a solution depend on perception versus abstract reasoning, and MathFlow decouples a dedicated perception module from a flexible inference LLM, showing that strengthening PAR’s Perception stage can improve performance across many backbones. Decoupled frameworks such as DVLR~\cite{guo2025integratingvisualinterpretationlinguistic} similarly separate visual interpretation from linguistic reasoning and adopt outcome-rewarded joint tuning on geometry benchmarks, while RL methods like VL‑Rethinker~\cite{wang2025vl} illustrate how, once process- and robustness-oriented benchmarks exist, self-reflective and perception-aware training strategies become natural responses. Viewed through PAR and APE, future benchmark design and model design should be co-planned: answer-only suites are still useful for breadth, but sustained progress will depend on more process-rich, dynamic, and executable benchmarks that expose failure modes at each PAR stage and support verifiable, visually grounded reasoning.

\section{Practical Design Guidelines}
\label{app:guidlines}
While our survey is organized along the PAR (Perception–Alignment–Reasoning) pipeline and the APE (Answer–Process–Executable) hierarchy, practitioners ultimately need concrete guidance on how to instantiate these abstractions in real systems. This subsection distills several practical design guidelines from the methods and benchmarks reviewed above and summarizes them as actionable take-home messages.
\paragraph{No universal optimal design.} A central observation of this survey is that there is no “one-size-fits-all” multimodal mathematical reasoner. Executable, symbol-heavy pipelines provide strong guarantees and debuggability but are fragile to noisy perception and expensive to annotate. In contrast, purely neural, latent pipelines offer flexibility and robustness to imperfect inputs, yet make it difficult to enforce or inspect the underlying mathematical structure. Similarly, always-on deep reasoning (e.g., search, RL, and intensive tool augmentation) can improve robustness on difficult instances, but may be unnecessary or even harmful for routine, low-stakes problems due to increased latency and potential overfitting to benchmark-specific reward signals.
\paragraph{Choosing APE Levels and Benchmarks.} For large-scale, low-stakes applications such as homework assistance or interactive practice, answer-level evaluation on broad suites like MathVista~\cite{lu2023mathvista} or ChartQA~\cite{masry2022chartqa} is often sufficient to guide model selection, provided that occasional errors are acceptable and qualitative inspection is used to detect obvious shortcut behavior. In safety-critical or high-stakes settings (e.g., automatic grading, high-level examinations, or formal theorem proving), process- or executable-level benchmarks—such as We-Math~\cite{qiao2024we}, MM-MATH~\cite{sun2024mm}, FlowVerse~\cite{chen2025mathflowenhancingperceptualflow}, NL2SQL-style datasets~\cite{zhong2017seq2sqlgeneratingstructuredqueries}, or formal geometry corpora—are preferable because they reveal where the reasoning chain fails and allow automatic verification of intermediate states. A practical rule of thumb is:
 (1) rely primarily on answer-level evaluation when coverage, scale, and latency are the dominant constraints and individual mistakes are tolerable;
 (2) adopt process-level evaluation when diagnosing typical failure modes (knowledge gaps, hallucinated steps, perception mistakes) is important;
 (3) favor executable-level evaluation when correctness and debuggability outweigh annotation cost and domain coverage.
\paragraph{Guidelines for Alignment Design.} When verifiability and fine-grained error analysis are paramount—for instance, in exam grading or systems that must provide legally or pedagogically reliable feedback—executable or DSL-based alignment (e.g., geometry DSLs, SQL, program-of-thought operators) combined with solver-backed checks is preferable, despite higher engineering and annotation overhead. For broad, latency-sensitive platforms such as large-scale tutoring systems, lightweight latent alignment with unified abstractions on top of generic MLLM backbones is often more appropriate, trading strict guarantees for robustness to noisy diagrams and lower maintenance cost. Hybrid designs that use executable alignment for a small set of core skills (e.g., Euclidean geometry, table/SQL reasoning) and latent alignment elsewhere provide a pragmatic compromise when both formal guarantees and wide task coverage are required.
\paragraph{Guidelines for Reasoning Paradigms.} For routine, low-stakes tasks, CoT-only or single-pass reasoning is typically adequate: such approaches are easy to deploy, respect strict latency budgets, and can be combined with simple calibration to reduce over-confident failures. For competition-level, research-style, or grading-style problems, RL-enhanced or search-based reasoning, often coupled with tool augmentation (e.g., calculators, theorem provers, program execution), is more suitable, as it prioritizes robustness and faithfulness over runtime. When both efficiency and reliability matter, selective or budgeted “think-more-when-needed” strategies form a practical middle ground: the model uses fast CoT for most inputs but automatically triggers deeper search or external tools on uncertain or adversarial cases, as indicated by uncertainty measures or self-consistency checks.
\paragraph{Recommended Configurations.} Putting these pieces together, several patterns emerge as practically useful design recipes:
 (1) Safety-critical grading and assessment: executable or DSL-based alignment with solver-backed checks, combined with search-based or RL-enhanced reasoning, evaluated predominantly at process or executable APE levels.
 (2) Large-scale tutoring and practice platforms: latent alignment with unified representations and fast CoT-style or shallow multi-step reasoning, primarily evaluated at the answer level, with spot checks on process-level benchmarks.
 (3) Interactive tools balancing guarantees and responsiveness: hybrid alignment (symbolic for a core subset of tasks, latent elsewhere) together with selective or budgeted multi-step or tool-augmented reasoning, evaluated with a mix of answer-, process-, and executable-level benchmarks.

\section{Systematic Failure Patterns in Practical Settings}
\label{app:failurepatterns}
While \autoref{tab:eval-pyramid} maps existing benchmarks to the Answer–Process–Executable (APE) hierarchy and the PAR stages, practical reliability also depends on how models fail in realistic conditions. Beyond aggregate scores, process-level, robustness, and comprehensive benchmarks expose recurring failure patterns that cut across perception, alignment, and reasoning. In this subsection, we synthesize these patterns along the PAR and APE dimensions, with a particular focus on sensitivity to low-quality diagrams, ambiguous multimodal references, and domain shifts between educational and scientific contexts.

\paragraph{Perception-level Failures.}Models exhibit sensitivity to low-quality diagrams, including low resolution, compression artifacts, cluttered layouts, partial crops, and imperfect OCR such as handwritten annotations. The manifestations are task-dependent: in geometry, small perturbations lead to missed intersections, distorted angles, or mis-detected primitives; in chart and table reasoning, they surface as axis, legend, and scale extraction errors; in visual math word problems, they obscure small objects or local relations. Robustness-oriented resources such as VCBench~\cite{li2024vcbench}, DynaMath~\cite{zou2024dynamath}, HC-M3D~\cite{liu2025rolevisualmodalitymultimodal}, and SMART-840~\cite{cherian2024evaluating} explicitly probe these sensitivities through multi-image dependencies, visual perturbations, and near-duplicate cases, while domain-specific sets like GeoMath and multi-image K–12 MV-MATH~\cite{zhougeomath,wang2025mv} add further perception stressors in scientific and educational contexts.
\paragraph{Alignment-level Failures.} A second class arises from ambiguous multimodal references and domain shifts between educational and scientific contexts. Errors include binding textual mentions such as “this triangle,” “the bar for 2021,” or “region A” to wrong regions, and unit or scale mismatches in charts and tables even when local perception is correct. Benchmarks such as ChartQA~\cite{masry2022chartqa}, PlotQA~\cite{methani2020plotqa}, FinQA~\cite{chen2022finqadatasetnumericalreasoning}, TAT-QA~\cite{zhu2021tatqaquestionansweringbenchmark}, ChartQAPro~\cite{masry-etal-2025-chartqapro}, and CharXiv~\cite{wang2024charxivchartinggapsrealistic} consistently reveal mistakes in axis and legend binding and unit normalization. Distributional differences between MM-K12~\cite{du2025mm}, OlympiadBench~\cite{he2024olympiadbench}, and Children’s Olympiads~\cite{cherian2024evaluating} versus scientific or photo-based suites such as MathScape~\cite{zhou2024mathscape} and SceMQA~\cite{liang2024scemqa} further cause executable descriptions that appear valid to encode wrong bindings or mismatched assumptions, reducing transfer across settings.
\paragraph{Reasoning-level Failures.} Even with mostly correct perception and alignment, models often produce unfaithful or brittle chains. Process-level evaluations show cases where models reach correct answers via unsupported steps, hallucinated operations not grounded in visuals, or sharp drops on out-of-distribution problems despite plausible narratives. Datasets such as MM-MATH~\cite{sun2024mm}, MPBench~\cite{xu2025mpbench}, ErrorRadar~\cite{yan2024errorradar}, Sherlock~\cite{ding2025sherlock}, We-Math~\cite{qiao2024we}, MathVerse~\cite{zhang2024mathverse}, CHAMP~\cite{mao2024champ}, and PolyMATH~\cite{gupta2024polymath} expose over-reliance on language priors, under-use of visual evidence, and gaps between answer-level success and process-level faithfulness. Executable resources including GeoQA+~\cite{cao2022augmented}, Geometry3K~\cite{lu2021intergpsinterpretablegeometryproblem}, E-GPS~\cite{wu2024gps}, and FormalGeo~\cite{zhang2024formalgeoextensibleformalizedframework} further reveal reasoning traces that fail strict program or proof checking despite coherent text, highlighting latent misalignments and logical inconsistencies.
\paragraph{Findings.} Viewed through PAR and APE, these patterns indicate that reliable deployment requires perception robust to degraded or stylistically varied diagrams, alignment that handles ambiguous references and cross-domain conventions including units and scales, and reasoning audited at the process and executable levels. Accordingly, evaluations should complement answer-level metrics with robustness suites, step-level diagnostics, and executable checks targeted to the failure modes most relevant to the application domain. We revisit these observations in Section~\ref{sec:challenges} and connect them to the task-specific failure modes summarized in Section~\ref{sec:perception}.

\begin{table*}[t]
\centering
\tiny
\setlength{\tabcolsep}{4pt}
\resizebox{0.98\textwidth}{!}{%
\begin{tabular}{l l l l l p{3.5cm} p{8.2cm}}
\toprule
\textbf{Task} & \textbf{Representative System} & \textbf{PAR} & \textbf{Highest APE} & \textbf{Primary Benchmarks} & \textbf{Executable Interface} & \textbf{Representative Performance} \\
\midrule

Geometry & GEOS & Perception + Alignment & Executable & GEOS (official \& practice SAT geometry) & Equation solver over parsed text + diagram & 49\% accuracy on official SAT geometry questions and 61\% on practice questions; on the $\sim$51\% of questions the system chooses to answer, accuracy exceeds 96\%. \\

Geometry & NGS / GeoQA program-supervised & Alignment + Reasoning & Executable & GeoQA, GeoQA+ & Program executor over symbolic programs & 60.0\% accuracy on GeoQA; the improved DPE-NGS reaches 62.65\% on GeoQA and 66.09\% on GeoQA+. \\

Geometry & Inter-GPS & Alignment + Reasoning & Executable & Geometry3K, GeoQA, GEOS & Geometry DSL / theorem rules & 78.3\% accuracy on Geometry3K and 68.0\% on GeoQA, clearly improving over earlier NGS (60\% on GeoQA); also outperforms GEOS on the GEOS dataset. \\

Geometry & PGPSNet & Alignment + Reasoning & Executable & Geometry3K, GeoQA & Program-supervised geometry DSL & 77.9\% accuracy on Geometry3K and 70.4\% on GeoQA. \\

Geometry & LANS & Alignment + Reasoning & Executable & Geometry3K, GeoQA & Geometry DSL with learned abstraction & 82.3\% accuracy on Geometry3K and 74.0\% on GeoQA, ranking among the strongest traditional GPS systems. \\

Geometry & FormalGeo-style provers / FGeo-HyperGNet & Alignment + Reasoning & Executable & FormalGeo7K & FormalGeo DSL + symbolic engine & Around 85.5\% overall accuracy and 87.7\% step-wise accuracy on FormalGeo7K, significantly outperforming previous geometry solvers. \\

Geometry & GeoDRL & Alignment + Reasoning & Executable & GeoQA & RL-guided theorem selection with symbolic solver & About 89.4\% accuracy on GeoQA, one of the highest reported results on this benchmark. \\

Geometry & Suffi-GPSC / FGeo-DRL series & Alignment + Reasoning & Executable & GeoQA, GeoQA+ & RL-guided formal solver & Suffi-GPSC achieves 87.4\% accuracy on GeoQA; FGeo-DRL reports 86.4\% on GeoQA, offering a trade-off between peak accuracy and proof interpretability. \\

Geometry & E-GPS & Perception + Alignment + Reasoning & Executable & Geometry3K, GeoQA & Top-down solver + bottom-up problem generator & Reports accuracy on Geometry3K and GeoQA comparable to Inter-GPS and GeoDRL, while substantially reducing average reasoning steps and improving interpretability (exact numbers are given in the original tables). \\

Geometry & Pi-GPS & Perception + Alignment + Reasoning & Executable & Geo170K, Geometry3K & Large-scale GPS pipeline with geometry DSL & Claims nearly a 10-point absolute improvement over previous neuro-symbolic GPS methods on Geometry3K and maintains state-of-the-art performance on the large-scale Geo170K corpus (exact percentages reported in the original paper). \\

Geometry & AlphaGeometry & Reasoning & Executable & IMO-AG (Olympiad geometry) & Formal theorem prover (DDAR) & Solves 25 of 30 recent IMO-AG geometry problems; the later AlphaGeometry2 variant solves 42 of 50 problems from 2000--2024 (84\% solve rate), surpassing the average human gold-medalist performance. \\

\midrule

Charts \& Tables & VisionTaPas & Alignment + Reasoning & Answer & ChartQA-H / ChartQA-M & Text + table encoder (non-pixel) & About 45.5\% overall accuracy on the original ChartQA test set, with 28.72\% on the harder ChartQA-H split and 53.84\% on ChartQA-M. \\

Charts \& Tables & Pix2Struct-Large & Perception + Alignment & Answer & ChartQA, AI2D, etc. & Fully visual encoder--decoder (no explicit table interface) & Achieves 58.6\% relaxed accuracy on ChartQA, improving the previous VisionTaPas result from 45.5\% to 58.6\%. \\

Charts \& Tables & ChartLlama & Perception + Alignment + Reasoning & Answer & ChartQA, chart-to-text, chart extraction & LLaVA-style VLM with chart-specific pre-training & Obtains 48.96\% accuracy on the original ChartQA test set and 90.36\% on the authors' ``special charts,'' for an average of 69.66\% across their two splits. \\

Charts \& Tables & ChartVLM & Perception + Alignment + Reasoning & Answer & ChartX (ChartQA-like multi-task benchmark) & Chart-specialized VLM & Around 40.71\% accuracy on the ChartQA-style task in ChartX, substantially outperforming general-purpose LMMs such as GPT-4V on this benchmark. \\

Charts \& Tables & GPT-4V / GPT-4o / LLaVA & Perception + Reasoning & Answer & ChartQA, ChartInsights & --- & GPT-4V/4o generally outperform open-source models such as InstructBLIP and LLaVA on chart reasoning; on the ChartInsights benchmark, GPT-4o reaches about 69.2\% accuracy, whereas the mean accuracy of 19 other open and closed models is only $\sim$39.8\%. \\

\midrule

VWP \& Mixed & LLaVA-13B & Perception + Reasoning & Answer & MathVista test & --- & Achieves 25.4\% overall accuracy on MathVista test, only modestly above the random baseline of 17.9\%. \\

VWP \& Mixed & CoT / PoT GPT-4 (caption + OCR tools) & Reasoning (tool-augmented) & Answer & MathVista test & External tools (image captioning + OCR) & CoT GPT-4 reaches 30.50\% accuracy and PoT GPT-4 reaches 31.74\% on MathVista, showing moderate gains from tool-augmented text-only pipelines. \\

VWP \& Mixed & GPT-4V & Perception + Reasoning & Answer & MathVista test & Direct image input & Achieves 49.9\% overall accuracy on MathVista test, about 15.1 points higher than Bard and still roughly 10.4 points below human performance (60.3\%). \\

VWP \& Mixed & Math-LLaVA-13B & Perception + Reasoning & Answer & MathVista testmini, MathVerse, etc. & --- & Reaches 46.6\% accuracy on MathVista testmini, improving over the LLaVA-1.5-13B base model by 19 absolute points and approaching GPT-4V on this split; also achieves competitive results on Math-V and related benchmarks. \\

\bottomrule
\end{tabular}%
}
\vspace{-0.05in}
\caption{Representative systems and reported performance on shared benchmarks.}
\vspace{-0.1in}
\label{tab:representative-systems-performance}
\end{table*}

\appendix

\end{document}